\documentclass[a4paper]{cas-dc}
\usepackage{bbm}

\usepackage[numbers]{natbib}

\def\tsc#1{\csdef{#1}{\textsc{\lowercase{#1}}\xspace}}
\tsc{WGM}
\tsc{QE}

\begin{document}
\let\WriteBookmarks\relax
\def\floatpagepagefraction{1}
\def\textpagefraction{.001}
\let\printorcid\relax

\shorttitle{Enhancing Point Annotations with Superpixel and Confident Learning Guided for Improving Semi-Supervised OCT Fluid Segmentation}    

\shortauthors{Tengjin Weng et al.}

\title [mode = title]{Enhancing Point Annotations with Superpixel and Confident Learning Guided for Improving Semi-Supervised OCT Fluid Segmentation}  



%


\author[1]{Tengjin Weng}

\author[2]{Yang Shen}
\cormark[1]

    \author[3]{Kai Jin}

\author[4]{Yaqi Wang}

\author[5]{Zhiming Cheng}

\author[6]{Yunxiang Li}

\author[2]{Gewen Zhang}

\author[7,8]{Shuai Wang}
\cormark[2]
\affiliation[1]{organization={School of Computer Science and Technology, Zhejiang Sci-Tech University},
                city={Hangzhou},
                postcode={310018}, 
                country={China}}
\affiliation[2]{organization={School of Engineering, Lishui University},
                city={Lishui},
                postcode={323000}, 
                country={China}}
\affiliation[3]{organization={Department of Ophthalmology, the Second Afﬁliated Hospital of Zhejiang University, College of Medicine},
                city={Hangzhou},
                postcode={310009}, 
                country={China}}
\affiliation[4]{organization={College of Media Engineering, Communication University of Zhejiang},
                city={Hangzhou},
                postcode={310018}, 
                country={China}}
\affiliation[5]{organization={School of Automation, Hangzhou Dianzi University},
                city={Hangzhou},
                postcode={310018}, 
                country={China}}
\affiliation[6]{organization={Department of Radiation Oncology, UT Southwestern Medical Center, Dallas},
                city={TX},
                postcode={75235}, 
                country={USA}}
\affiliation[7]{organization={School of Cyberspace, Hangzhou Dianzi University},
                city={Hangzhou},
                postcode={310018}, 
                country={China}}

\affiliation[8]{organization={Suzhou Research Institute of Shandong University},
                city={Suzhou},
                postcode={215123}, 
                country={China}}

\cortext[1]{Corresponding author at: School of Engineering, Lishui University, Lishui 323000, China. Email address: tlsheny@163.com. (Y. Shen)}
\cortext[2]{Corresponding author at: School of Cyberspace, Hangzhou Dianzi University, Hangzhou 310018, China and also Suzhou Research Institute of Shandong University, Suzhou 215123, China. Email address: shuaiwang.tai@gmail.com. (S. Wang)}


\begin{abstract}
Automatic segmentation of fluid in Optical Coherence Tomography (OCT) images is beneficial for ophthalmologists to make an accurate diagnosis. Although semi-supervised OCT fluid segmentation networks enhance their performance by introducing additional unlabeled data, the performance enhancement is limited.
To address this, we propose Superpixel and Confident Learning Guide Point Annotations Network (SCLGPA-Net) based on the teacher-student architecture, which can learn OCT fluid segmentation from limited fully-annotated data and abundant point-annotated data.
Specifically, we use points to annotate fluid regions in unlabeled OCT images and the Superpixel-Guided Pseudo-Label Generation (SGPLG) module generates pseudo-labels and pixel-level label trust maps from the point annotations. The label trust maps provide an indication of the reliability of the pseudo-labels. Furthermore, we propose the Confident Learning Guided Label Refinement (CLGLR) module identifies error information in the pseudo-labels and leads to further refinement.
Experiments on the RETOUCH dataset show that we are able to reduce the need for fully-annotated data by 94.22\%, closing the gap with the best fully supervised baselines to a mean IoU of only 2\%.
Furthermore, We constructed a private 2D OCT fluid segmentation dataset for evaluation. Compared with other methods, comprehensive experimental results demonstrate that the proposed method can achieve excellent performance in OCT fluid segmentation.
\end{abstract}


\maketitle
\section{Introduction}
\label{sec:introduction}
Optical coherence tomography (OCT)~\cite{huang1991optical} is a powerful imaging technique used to acquire structural and molecular information of biological tissues.
OCT can noninvasively reconstruct high-resolution cross-sectional images from the backscattered light spectrum of biological samples~\cite{medeiros2005evaluation} by employing low-coherence interferometry.
It has found wide applications in various fields, particularly in biomedical applications such as ophthalmic imaging, cardiovascular imaging, gastrointestinal imaging, and pulmonary imaging. In traditional excision biopsies, the image quality obtained from within the stomach is often low, whereas endoscopic OCT can clearly depict the tissue microstructure inside the gastrointestinal tract.
Notably, ophthalmology stands as one of the earliest adopters of OCT technology~\cite{huang1991optical,trichonas2014optical}, ushering fundamental transformations in clinical practices within the field.

Macular edema (ME) is caused by the breakdown of the blood-retinal barrier leading to fluid infiltration in the macular area~\cite{marmor2000mechanisms} and is associated with retinal diseases such as age-related macular degeneration (AMD) and diabetic macular edema (DME)~\cite{hu2019automated}.
Research has shown a strong correlation between OCT signals and retinal histology, which is highly valuable for diagnosing ME caused by various diseases.
The total retinal thickness measured from OCT images is widely utilized for the diagnosis of ME, and numerous methods for layer segmentation in OCT images have been proposed~\cite{liu2019shortest,roy2017relaynet}.
However, studies have indicated that retinal fluid volume can provide a more accurate indication of vascular permeability~\cite{waldstein2016correlation}.
Retinal fluid can be divided into three types according to the accumulation site, which are subretinal fluid (SRF), intraretinal fluid (IRF), and pigment epithelial detachment (PED). Understanding the presence and location of these fluids can help ophthalmologists diagnose and monitor these conditions and develop appropriate treatment plans to preserve vision. Ophthalmologists rely on OCT images to identify the type and size of the fluid area, but accurately quantifying the fluid and formulating treatment plans can be challenging. Automated and accurate quantification of fluids in OCT images can significantly improve the efficiency of ophthalmologists' diagnosis and treatment. 

Traditional segmentation techniques, such as threshold-based \cite{wilkins2012automated}, graph-based \cite{wang2016label, rashno2017fully}, and machine learning-based \cite{montuoro2017joint} methods, have been utilized for OCT segmentation in the past, but they are often susceptible to variations in image quality, require extensive domain knowledge, and lack generalization capabilities.

In contrast to traditional segmentation methods that rely on carefully crafted handcrafted features, convolutional neural networks (CNNs) can automatically learn and extract image features from the data itself. Therefore, various CNN-based methods have been developed for performing segmentation tasks, such as FCN~\cite{long2015fully}, SegNet~\cite{badrinarayanan2017segnet}, DeepLab~\cite{chen2017deeplab}, and UNet~\cite{ronneberger2015u}. The utilization of CNNs in medical image segmentation requires substantial amounts of data. Unfortunately, manual segmentation of medical images demands significant expertise and time. Obtaining an adequate quantity of accurately labeled data from medical experts can be a difficult and challenging task, thereby posing obstacles to developing precise CNN models for medical image segmentation. 

To address these limitations, Semi-Supervised Learning (SSL) has garnered significant attention in medical image segmentation. The exploration of available unlabeled data is of great value for training segmentation models. SSL methods effectively leverage unlabeled data to improve the performance of segmentation models while reducing the reliance on labeled data. However, unlabeled data tends to have a limited improvement in the model's performance. Therefore, augmenting unlabeled data with additional weak annotations has been shown to be beneficial~\cite{dong2018learning}. These annotations, while not as detailed as fully-annotated data, still convey valuable information to the segmentation models. The process enhances the model's understanding of the data and improves its ability to generalize across diverse cases. It effectively bridges the gap between fully labeled and completely unlabeled data, contributing to even better segmentation performance. 
Point annotations require only a click on each category in the image, which is easily obtained as labels in OCT fluid segmentation.
Therefore, we choose to perform point annotation on unlabeled data in semi-supervised OCT fluid segmentation to minimize human effort.

However, the point annotations only cover a tiny area within the fluid region, lacking any extension, shape, or boundary information of the fluid, which results in point annotations not providing sufficient information about the fluid for training. To address this, a popular weakly supervised learning approach is to generate pseudo-labels from weak point annotations and then use these pseudo-labels to train segmentation models. Nevertheless, this method may result in degraded model performance because the generated pseudo-labels may be imprecise, and using these pseudo-labels to train segmentation models may introduce errors.

In this work, we propose Superpixel and Confident Learning Guide Point Annotations Network (SCLGPA-Net) constructed by a teacher-student architecture to reduce the reliance on fully-annotated data in automated OCT fluid segmentation. The detailed architecture of the proposed network framework is depicted in the upper part of Fig~\ref{fig1}.
Firstly, our approach is in semi-supervised mode and performs simple point annotations on unlabeled data. We introduce the Superpixel-Guided Pseudo-Label Generation (SGPLG) module to generate pseudo-labels and label trust maps (weight maps) based on point annotations. These pseudo-labels with weights are subsequently utilized in the training process, enabling the development of the Superpixel-Guided Point Annotations Network (SGPA-Net). Secondly, we introduce the Confident Learning Guided Label Refinement (CLGLR) module to identify and refine error labels in pseudo-label in conjunction with the SGPA-Net predictions, resulting in refined-label and getting the SCLGPA-Net with better performance. 
The contributions of our research are summarized as follows:

\begin{itemize}
\item[$\bullet$] We propose a semi-supervised SCLGPA-Net, constructed by a teacher-student architecture. By leveraging additional point annotations to enrich the unlabeled data to enhance the pure image information provided into more valuable weakly supervised information. This conversion can significantly boost model performance while reducing the need for precise annotations.

\item[$\bullet$] We propose SGPLG, a superpixel-guided method for generating pseudo-labels and label trust maps based on point annotation. The label trust maps constrain the network from fitting label errors by assigning lower confidence to suspected noisy label pixels.
 
\item[$\bullet$] To our best knowledge, this is the first research that applies the label-denoising method to OCT fluid segmentation. The proposed CLGLR can identify labeling errors in the pseudo-labels through confidence calibration under the constraints of the label trust map and perform further refinement to obtain more accurate labels to further improve model performance. 
\end{itemize}

\section{Related Work}
\subsection{CNN-Based OCT Fluid Segmentation}
Many successful OCT fluid segmentation methods use convolutional neural networks (CNNs) based on the UNet~\cite{ronneberger2015u} architecture. Rashno~\textit{et al.} ~\cite{rashno2018oct} incorporated a graph shortest path technique as a post-processing step to enhance the predictive results of UNet for OCT fluid segmentation. To exploit the structural relationship between retinal layers and fluids, Xu~\textit{et al.}~\cite{xu2017dual} proposed a two-stage fluid segmentation framework. They first trained a retinal layer segmentation network to extract retinal layer maps which were used to constrain the fluid segmentation network in the second stage. Several other studies, such as~\cite{hassan2019deep,lu2019deep}, employed a graph-cut method to generate retinal layer segmentation maps. These maps were then combined to train a UNet for fluid segmentation. Moreover, De~\textit{et al.}~\cite{de2018clinically} proposed a UNet-based architecture that can simultaneously segment retinal layers and fluids, utilizing pixel-level annotations of retinal layer and fluid masks to enhance OCT segmentation performance.

Although various methods have been proposed with little difference in performance, the effectiveness of current OCT fluid segmentation methods relies heavily on a large number of datasets with precision annotations. Notably, obtaining a large amount of precision annotation for these images is often challenging. For this reason, semi-supervised methods that explore the available unlabeled data are a reliable solution.

\subsection{Semi-Supervised Segmentation}
Considerable efforts have been devoted to advancing semi-supervised medical image segmentation. Among them, consistency regularization has been widely used in semi-supervised segmentation.

A classic framework is Mean-Teacher (MT) ~\cite{tarvainen2017mean} and researchers have extended the Mean-Teacher framework in various ways to enhance its capabilities. For instance, UA-MT~\cite{yu2019uncertainty} incorporates uncertainty information to guide the student network in gradually learning from reliable and meaningful targets provided by the teacher network. SASSNet~\cite{li2020shape} leverages unlabeled data to enforce geometric shape constraints on segmentation results. DTC~\cite{luo2021semi} introduces a dual-task consistency framework, explicitly incorporating task-level regularization. BCP~\cite{bai2023bidirectional} introduces a bidirectional CutMix~\cite{yun2019cutmix} approach to facilitate comprehensive learning of common semantics from labeled and unlabeled data in both inward and outward directions.
Other methods, like ICT~\cite{verma2022interpolation}, encourage coherence between predictions at interpolated unlabeled points and the interpolation of predictions at those points. CPS~\cite{chen2021semi} employs two networks with identical structures but different initializations, imposing constraints to ensure their outputs for the same sample are similar. These innovative approaches collectively contribute to further enhancing the effectiveness of semi-supervised medical image segmentation. 

Furthermore, several studies \cite{ji2019invariant,perone2019unsupervised,liu2022uncertainty,reiss2021every,seibold2022reference,wang2023deep} have specifically delved into OCT fluid segmentation under the semi-supervised setting. Liu~\textit{et al.}~\cite{liu2022uncertainty}  proposed UGNet in their research, which is a semi-supervised learning model guided by uncertainty for retinal fluid segmentation in OCT images. Reiss~\textit{et al.}~\cite{reiss2021every} proposed a novel segmentation training mechanism in their research, which can be flexibly used for OCT fluid segmentation. It allows the utilization of different types of annotated data during the training process, including fully labeled images, images with bounding boxes, only global labels, or even no annotated images at all. Seibold~\textit{et al.}~\cite{seibold2022reference} argue that visually similar regions between labeled and unlabeled images may potentially share the same semantics and, therefore, should share their labels. Following this approach, they use a small number of labeled images as reference material and match pixels in an unlabeled image to the semantics of the best-fitting pixel in a reference set.

However, unlabeled data can only provide image-only information to the model, resulting in limited improvement of model performance. In addition, enhancing unlabeled data with additional weak annotations has been demonstrated to be beneficial. While these annotations may not be as detailed as fully-annotated data, they still convey valuable information to the segmentation model and can significantly enhance model performance.
 
\subsection{Weakly-Supervised Segmentation}

Several researchers have developed some weakly supervised OCT fluid segmentation methods. He~\textit{et al.}~\cite{he2022intra} introduced a method dubbed Intra-Slice Contrast Learning Network (ISCLNet) that relies on weak point supervision for 3D OCT fluid segmentation. However, in actual diagnoses, ophthalmologists typically only concentrate on a limited number of OCT images displaying fluid. The inter-image comparison technique deployed by ISCLNet can be challenging when dealing with incomplete OCT data. 

In medical image segmentation, a prevalent approach of weakly-supervised learning involves generating pseudo-labels from weak annotations and subsequently using these to train a segmentation model. Pu~\textit{et al.}~\cite{pu2018graphnet} proposed a technique that utilizes a graph neural network based on superpixels to create pseudo-labels from weak annotations like points or scribbles. Nevertheless, this method could introduce two sources of error: inaccuracies in the generated pseudo-labels and the subsequent errors in learning segmentation from these labels. 
To address this challenge, a reliable approach is to utilize weighted pseudo-labels to mitigate the interference of erroneous labels during training. Another method involves employing label-denoising techniques to filter or reduce noise within these labels effectively.

\subsection{Learning Segmentation with Noisy Labels}
Previous work has pointed out that labeled data with noise can mislead network training and degrade network performance. Most existing noise-supervised learning works focus on image-level classification tasks~\cite{northcutt2021confident},~\cite{han2018co},~\cite{xue2019robust} while more challenging pixel-wise segmentation tasks remain to be studied. Zhang~\textit{et al.} ~\cite{zhang2020robust} proposed a TriNet based on Co-teaching~\cite{han2018co}, which trains a third network using combined predictions from the first two networks to alleviate the misleading problem caused by label noise. Zhang~\textit{et al.}~\cite{zhang2020characterizing} suggested a two-stage strategy for pre-training a network using a combination of different datasets, followed by fine-tuning the labels by Confident Learning to train a second network. Zhu~\textit{et al.}~\cite{zhu2019pick} proposed a module for assessing the quality of image-level labels to identify high-quality labels for fine-tuning a network. Xu~\textit{et al.}~\cite{xu2022anti} developed the MTCL framework based on Mean-Teacher architecture and Confident Learning, which can robustly learn segmentation from limited high-quality labeled data and abundant low-quality labeled data. The KDEM~\cite{dolz2021teach} method is an extension of the semi-supervised learning approach proposed by~\cite{luo2020semi}, which introduces additional techniques such as knowledge distillation and entropy minimization regularization to further improve the segmentation performance. Yang~\textit{et al.}~\cite{yang2022learning} introduce a dual-branch network that can learn efficiently by processing accurate and noisy annotations separately. 

These methods demonstrate how to improve the network's ability to learn noisy labels and provide insights for future research in this area. Extensive experiments of many label-denoising methods on datasets such as JSRT~\cite{shiraishi2000development} and ISIC~\cite{codella2018skin} have achieved promising results, but limitations caused by the lack of OCT fluid segmentation datasets hinder the application of these methods. Therefore, the effectiveness of label-denoising methods in OCT fluid segmentation remains largely unexplored.

\begin{figure*}
    \centerline{\includegraphics[width=\textwidth,height=1\textwidth]{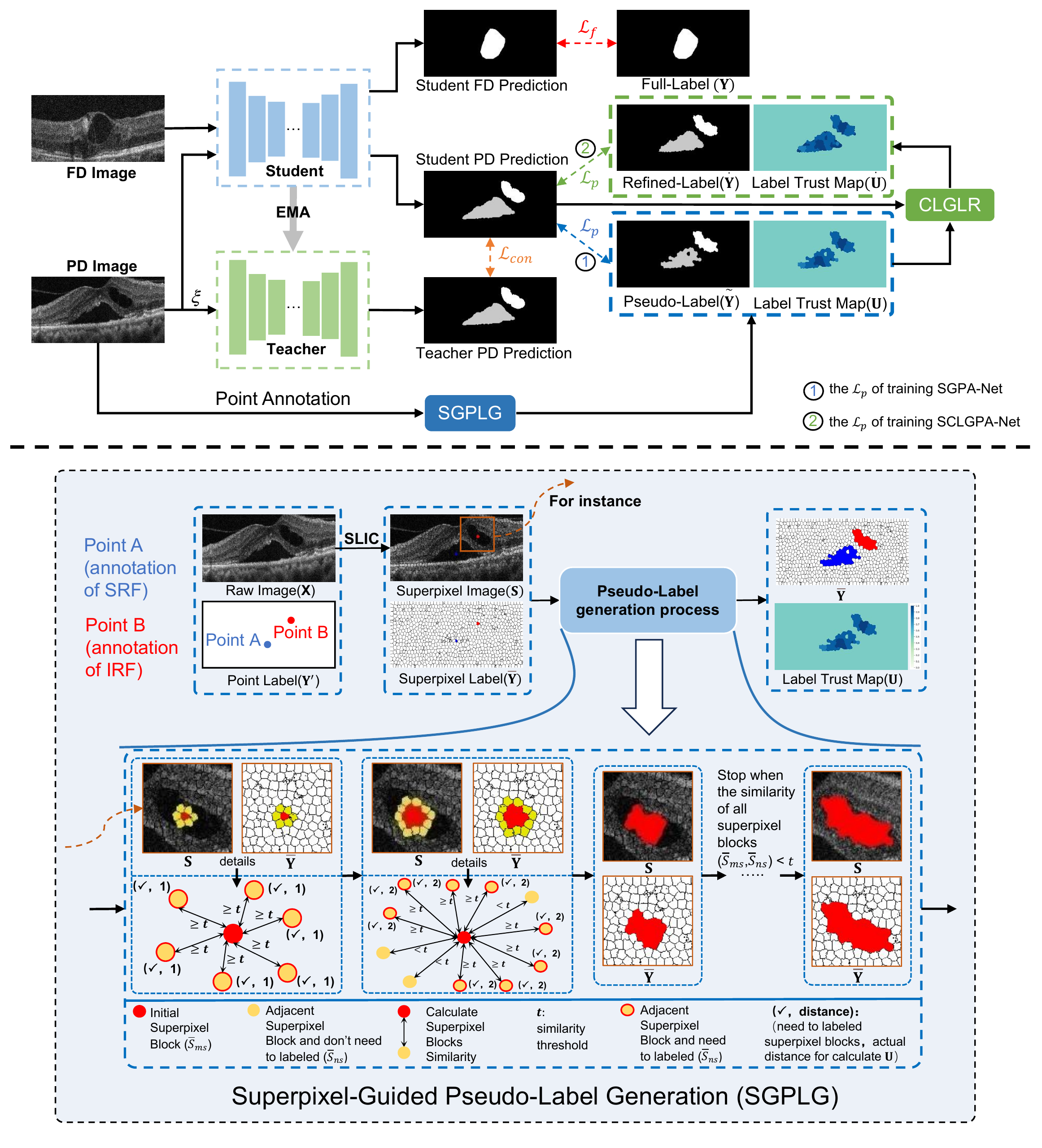}}
    \caption{Illustration of SCLGPA-Net. The top part provides an overall of our approach, while the bottom part focuses on the Superpixel-Guided Pseudo-Label Generation (SGPLG) module. The details of the Confident Learning Guided Label Refinement (CLGLR) module are shown in Fig.~\ref{fig2}. The images of the FD are fed to the student model, and the images of the PD are both fed to the student model and teacher model. Simultaneously, the SGPLG generates pseudo-labels and label trust maps of PD. After obtaining the student network (SGPA-Net) based on teacher-student architecture, CLGLR is used to identify label errors within the pseudo-label, and the estimated error map is obtained to guide refining labels. Update the original pseudo-label with the refined-label, and train the same network with the teacher-student architecture again to obtain SCLGPA-Net.
}
    \label{fig1}
\end{figure*}
\begin{figure}[htb]
\centerline{\includegraphics[width=\linewidth]{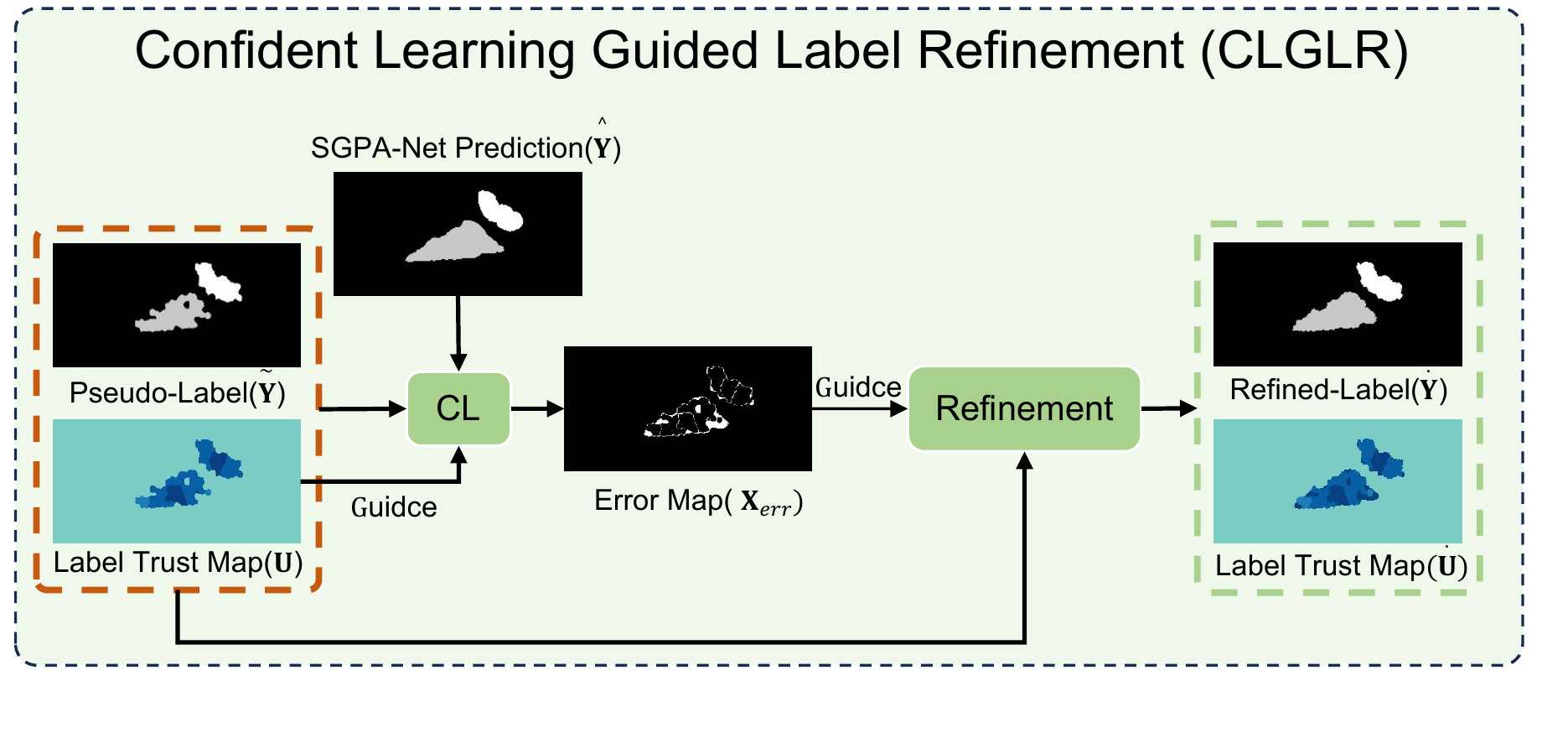}}
    \caption{Illustration of the Confident Learning Guided Label Refinement (CLGLR) module.}
    \label{fig2}
\end{figure}

\section{Methodology}

\subsection{Framework Overview}
Our method divides the dataset into two groups: fully-annotated data (FD) and point-annotated data (PD). To simplify the description of our methodology, we define $M$ samples to represent the FD, while the remaining $N - M$ samples represent the PD. We denote the FD as $\mathbf{D}_f=\{( \mathbf X_{(i)},  \mathbf Y_{(i)})\}^M_{i=1}$ and the PD as $\mathbf{D}_p=\{( \mathbf X_{(i)}, \mathbf{Y}'\}^N_{i=M+1}$, where $ \mathbf X_{(i)} \in \mathbb{R}^{\Omega _i}$ represents the input 2D OCT images. $ \mathbf Y_{(i)}, \mathbf{Y}'_{(i)}  \in  \{0,1,2,3\}^{\Omega _i}$ (four types of segmentation tasks) denotes the full-label and point-label of OCT image, respectively.


Fig.~\ref{fig1} illustrates SCLGPA-Net that aims to learn OCT fluid segmentation simultaneously from limited FD and abundant PD. The images of the FD are fed to the student model, and the images of the PD are both fed to the student model and teacher model. Simultaneously, the SGPLG module generates pseudo-labels and label trust maps of PD. After obtaining the student network (SGPA-Net) based on teacher-student architecture, CLGLR is used to identify label errors within the pseudo-label, and the estimated error map is obtained to guide refining labels. Update the original pseudo-label with the refined-label, and train the same network with the teacher-student architecture again to obtain SCLGPA-Net. More details about the framework of SCLGPA-Net are explained in the following.

\subsection{SGPLG for Generating Pseudo-labels and Label Trust Maps}

Our proposed SGPLG can generate pseudo-labels and label trust maps from weak point annotations via superpixel guidance. The pixel-level label trust maps provide an indication of the reliability of these pseudo-labels.

\subsubsection{Superpixel-Guided for Generating Pseudo-Labels from Point Annotations}

Superpixel segmentation algorithms, such as SLIC~\cite{achanta2012slic},  LSC~\cite{li2015superpixel}, and Manifold-SLIC~\cite{liu2016manifold} calculate feature similarities (e.g., color, brightness, texture, and shape) among adjacent pixels in an image. They subsequently group these pixels into visually meaningful entities. Consequently, they preserve the boundary information of objects within the image, while significantly reducing the complexity of subsequent image-processing tasks. We consider integrating the superpixel-guided strategy into OCT fluid segmentation, as it aids in the development of robust and efficient diagnostic tools. 

The segmentation targets for the OCT fluid segmentation dataset include three types of fluids: PED, SRF, and IRF. For point labeling, we make a single marker point in the region of the SRF and IRF and multiple points (connected to a line) in the bottom region of the PED. Note that all points passing through the line when labeling the PED are point annotations.

Formally, give a raw image $\mathbf{X}$, the point label is represented by $\mathbf{Y}' = \{Y'_i\}_{i=1}^n$, $Y'_i \in \{1,2...,C\}$ where $C$ is the number of semantic classes and $n$ is the number of pixels. The superpixel image is obtained based on the SLIC~\cite{achanta2012slic} algorithm. We denote superpixel image as $\mathbf{S} = \{S_i\}_{i=1}^n$, where $S_i \in \{1,2...,K\}$ and the $K$ is the number of superpixel blocks. Here $S_j = k$ means that the pixel $j$ belongs to the $k^{th}$ superpixel block. We can represent all the pixels $j$ that are included in the $k^{th}$ superpixel block by $\mathbf{\bar{S}} = \{\bar{S}_k\}_{k=1}^{K}$, where $ \bar{S}_k = \{j \mid S_{j = k}\}_{j=0}^{n}$. Further, the superpixel label is represented by $\mathbf{\bar{Y}} = \{\bar{Y}_k\}_{k=1}^{K}$ and the initial values are zero (background). 

The following procedure illustrates how to convert point-label $\mathbf{Y}'$ to superpixel label $\mathbf{\bar{Y}}$:
\begin{equation}
\bar{Y}_k = c,    \exists\ (Y'_j = c), 
\end{equation}
where $j \in \bar{S}_k$ and $c \ne 0$. From this, we get the initial superpixel label $\mathbf{\bar{Y}}$. Next, we iterate over the initial superpixel label based on the similarity of the superpixel blocks to obtain the final superpixel label (the pseudo-label generation process in the figure of SGPLG).

Due to a scarcity of pixel annotations in $\mathbf{Y}'$, the majority of $\bar{Y}_k$ values are equal to zero. Identify and isolate all $\bar{Y}_k$ not equal to 0 and randomly select a superpixel block label $\bar{Y}_{ms}$. The corresponding superpixel block is $\bar{S}_{ms}$. We select one of the adjacent (all adjacent superpixel blocks for IRF and SRF, upper adjacent superpixel block for PED) superpixel blocks $\bar{S}_{ns}$ of $\bar{S}_{ms}$ and performs the following operations:
\begin{equation}
    \bar{Y}_{ns} = \bar{Y}_{ms} \cdot \mathbbm{I}(cos\_dis(\bar{S}_{ms}, \bar{S}_{ns}) \ge t),
\end{equation}
where $cos\_dis(\bar{S}_{ms}, \bar{S}_{ns})$ represents the superpixel blocks similarity and $t$ is the similarity threshold (the similarity threshold of IRF and SRF to 0.6, and the similarity threshold of PED to 0.5). The similarity of  $\bar{S}_{ms}$ and $\bar{S}_{ns}$ as follows:
\begin{equation}
    cos\_dis(\bar{S}_{ms},\bar{S}_{ns})=\frac{\sum_{v=0}^{255}{(\mathbf{O}_{ms}^v)(\mathbf{O}_{ns}^v)}}{\sqrt{\sum_{v=0}^{255}{(\mathbf{O}_{ms}^v)^2}}\sqrt{\sum_{v=0}^{255}{(\mathbf{O}_{ns}^v)^2}}}, \\
\end{equation}
where $\mathbf{O}_k^v$ represents the number of pixel values $v$ contained in the $k^{th}$ superpixel block. The number of each pixel value contained in the $\bar{S}_{ms}$ and $\bar{S}_{ns}$ are calculated by the following formula:
\begin{equation}
\begin{split}
    &\mathbf{O}_{ms}^v = \sum_{j:S_j=ms} \mathbbm{I}(\mathbf{X}[j] = v ), v=[0,255], \\
    &\mathbf{O}_{ns}^v = \sum_{j:S_j=ns} \mathbbm{I}(\mathbf{X}[j] = v ),v=[0,255].\\
\end{split}
\end{equation}

If $cos\_dis(\bar{S}_{ms},\bar{S}_{ns}) \ge t$, we assign the value of $\bar{Y}_{ms}$ to $\bar{Y}_{ns}$ and it is obvious that the adjacent superpixel blocks of $\bar{Y}_{ns}$ are also very likely to be similar to $\bar{Y}_{ms}$. Therefore, the adjacent superpixel blocks of $\bar{S}_{ns}$ will be regarded as the adjacent superpixel blocks of $\bar{S}_{ms}$. The processing of $\bar{Y}_{ms}$ will not end until all the similarity values of adjacent superpixel blocks are less than threshold $t$. After processing all initial $\bar{Y}_k$ not equal to 0, the final superpixel label $\mathbf{\bar{Y}}$ will be converted to the pseudo-label $\mathbf{\tilde{Y}} = \{\tilde{Y}\}_{i=1}^{n}$:
\begin{equation}
    \tilde{Y}_i = \bar{Y}_{S_i}.
\end{equation}

In the process of generating pseudo-labels, it is unreasonable to give the same confidence to all labels. Therefore, we propose a method to assign suitable confidence by measuring the actual distance of superpixel blocks.
Specifically, we introduce a pixel-wise label trust map $\mathbf{U} = \{U_i\}_{i=1}^n$ where $U_i \subseteq (0, 0.1, \ldots, 1)$. The label trust map is used to adjust the influence of each pixel's label during training, which can help mitigate the impact of misinformation on the network. All values of $U_i$ are set to 0.5 (the $U_i$ value of the pixels contained in the initial $\bar{Y}_k$ not equal to 0 is set to 1) and update $\mathbf{U}$ at the same time when updating $\mathbf{\bar{Y}}$. If $cos\_dis(\bar{S}_{ms},\bar{S}_{ns}) \ge t$, calculate the superpixel block distance between $\bar{S}_{ms}$ and $\bar{S}_{ns}$ and assign lower confidence values to the corresponding pixels on $\mathbf{U}$ that are farther away from the $\bar{S}_{ms}$. The specific value allocation formula is as follows:
\begin{equation}
    value = max(1.0 - 0.1 * \frac{distance}{2}, 0.0).
\end{equation}

\subsection{Training SGPA-Net Based on Mean-Teacher Architecture}
After being processed by the SGPLG module, $\mathbf{D}_p=\{( \mathbf X_{(i)}, \mathbf{Y}'\}^N_{i=M+1}$ change to $\mathbf{D}_p=\{( \mathbf X_{(i)},  \mathbf{\tilde{Y}}_{(i)},\mathbf{U}_{(i)})\}^N_{i=M+1}$.
The basic network architecture chooses the Mean-Teacher (MT) model, which is effective in SSL. The MT architecture comprises a student model (updated through back-propagation) and a teacher model (updated based on the weights of the student model at different training stages). A great strength of the MT framework is superior in its ability to leverage knowledge from image-only data using perturbation-based consistency regularization. 

Formally, we denoted the weights of the student model at training step $t$ as $\theta_t$. We updated the teacher model's weights $\widetilde{\theta}_t$ using an exponential moving average (EMA) strategy, which can be formulated as follows:

\begin{equation}
\widetilde{\theta}_t = \alpha\widetilde{\theta}_{t-1} + (1-\alpha)\theta_t,
\end{equation}
where $\alpha$ is the EMA decay rate, and it is set to 0.99, as recommended by~\cite{tarvainen2017mean}. Based on the smoothness assumption~\cite{luo2018smooth}, we encouraged the teacher model’s temporal ensemble prediction to be consistent with that of the student model under different perturbations, such as adding random Gaussian noise $\xi$ to the input images. Based on the MT architecture, we add pseudo-labels to unlabeled data. The student model learns from three aspects: constraints from full-label, constraints from pseudo-labels, and constraints from the consistency of the teacher model. The student network in the architecture after training is SGPA-Net.

\subsection{Confident Learning for Multi-Category Pixel-Wise Conditional Label Errors}
Despite the presence of label trust maps $\mathbf{U}$, which are designed to limit the impact of incorrect noise information on model learning, there remains the potential for noise information to be learned by the model. Confident Learning (CL)~\cite{northcutt2021confident} is able to identify label errors in datasets and enhance training with pseudo-labels by estimating the joint distribution between the noisy (observed) labels $\tilde{y}$ and the true (latent) labels $y^*$, as assumed by Angluin~\cite{angluin1988learning}. This estimation enables CL to assign higher confidence to instances with more reliable labels and lower confidence to instances with more questionable labels, resulting in finding the error labels. Zhang~\textit{et al.}~\cite{zhang2020characterizing} pioneered the application of CL to medical image segmentation and achieved promising results. Moreover, many follow-up studies~\cite{xu2021noisy},~\cite{xu2022anti} have proved the effectiveness of CL for medical image segmentation. However, most of the research is based on the segmentation task of binary classification, further research is needed to explore the effectiveness of CL for multi-category medical image segmentation.

Specifically, given PD image $\mathbf X$  and we denote $ \mathbf X = (\mathbf x, \tilde{y})^n$, where $\tilde{y}$ means the label of pixels and $n = w \times h$ means the number of pixels in $\mathbf{X}$. We can obtain the predicted probabilities $\hat{\mathbf{P}}$ for $m$ classes by the SGPA-Net. Assuming that a pixel $\mathbf{x}$ labeled $\tilde{y} = i$ has large enough predicted probabilities $\hat{\mathbf{P}}_j(\mathbf{x}) \ge t_j$, there is a possibility that the current annotation for $\mathbf{x}$ is incorrect, and it may actually belong to the true latent label $y^* = j$ ($ i \in \mathcal{C}_m$, $ j \in \mathcal{C}_m$, $\mathcal{C}_m$ indicates the set of $m$ class label). Here, we set the average predicted probabilities $\hat{\mathbf{P}}_j(\mathbf{x})$ of all pixels labeled $\tilde{y} = j$ as the threshold $t_j$:
\begin{equation}
t_j := \frac{1}{\left | \mathbf{X}_{\tilde{y} = j} \right | } \sum_{\mathbf{x} \in \mathbf{X}_{\tilde{y} = j}} {\hat{\mathbf{P}}_j(\mathbf{x}}).
 \end{equation}

Based on this assumption, we can construct the confusion matrix $\mathbf{C}_{\tilde{y}, y^*}$ by counting the number of pixels $\mathbf{x}$ that are labeled as $\tilde{y} = i$ and may actually belong to the true latent label $y^* = j$. The $\mathbf{C}_{\tilde{y}, y^*}[i][j]$ represents the count of such pixels for which the observed label is $\tilde{y} = i$ and the true latent label is $y^* = j$:

\begin{equation}
\begin{aligned}
\mathbf{C}_{\tilde{y}, y^*}[i][j] := \left | \hat{\mathbf{X}}_{\tilde {y}=i,y^*=j}  \right |,
\end{aligned}
\end{equation}
where
\begin{equation}
\begin{aligned}
 \hat{\mathbf{X}}_{\tilde {y}=i,y^*=j} := \{\mathbf{x} \in  \mathbf{X}_{\tilde {y}=i}:&\hat{\mathbf{P}}_j(\mathbf{x})\ge t_j, \\ & j=\mathop{\arg\max}\limits_{k \in M:\hat{\mathbf{P}}_k(\mathbf{x})\ge t_k }\hat {\mathbf{P}}_k(\mathbf{x} ) \}.
 \end{aligned}
\end{equation}

After obtaining the confusion matrix $\mathbf{C}_{\tilde{y}, y^*}$ it needs to be normalized. Then, the joint distribution $\mathbf{Q}_{\tilde{y}, y^*}$ between the pseudo-labels and the true labels can be obtained by dividing each element in the confusion matrix by the total number of pixels:
\begin{equation}
\begin{aligned}
\mathbf{Q}_{\tilde{y}, y^*}[i][j]= \frac{\mathbf{\tilde{C}}_{\tilde{y}, y^*}[i][j]}{ {\textstyle \sum_{i \in \mathcal{C}_m, j \in \mathcal{C}_m}{\mathbf{\tilde{C}}_{\tilde{y}, y^*}[i][j]}}},
\end{aligned}
\end{equation}
where 
\begin{equation}
\begin{aligned}
\mathbf{\tilde{C}}_{\tilde{y}, y^*}[i][j]=\frac{\mathbf{C}_{\tilde{y}, y^*[i][j]}}{ {\textstyle \sum_{j \in \mathcal{C}_m}{\mathbf{C}_{\tilde{y}, y^*}[i][j]}} 
} \cdot \left | \mathbf{X}_{\tilde{y}=i}  \right |.
\end{aligned}
\end{equation}

\begin{table*}[!t]
\centering
    \renewcommand\arraystretch{1.5}
    \caption{Oct Fluid Segmentation Studies on Private Dataset. Comparison of the Experimental Results of Other Semi-Supervised Methods on Different Ratios of FD and PD. The Best Results are in Bold. (Dice Unit: $\%$)}
    \label{tab1}
    \resizebox{0.78\linewidth}{!}{
\begin{tabular}{c|ccc|cccc}
\toprule[1pt]
\hline
\multirow{2}{*}{\textbf{Methods}} & \multicolumn{3}{c|}{\textbf{Settings}}                                & \multicolumn{4}{c}{\textbf{Metrics}}                                                                                                                           \\ \cline{2-8} 
                         & \multicolumn{1}{c|}{FD}    & \multicolumn{1}{c|}{PD}    & Point$\_$Annotation        & \multicolumn{1}{c|}{DSC}      & \multicolumn{1}{c|}{D$_{SRF}$}      & \multicolumn{1}{c|}{D$_{IRF}$}      & \multicolumn{1}{c}{D$_{PED}$}\\ \hline\hline
FD-Sup                   & \multicolumn{1}{c|}{20\%} & \multicolumn{1}{c|}{0\%} & $\times$    & \multicolumn{1}{c|}{73.54} & \multicolumn{1}{c|}{81.93} & \multicolumn{1}{c|}{76.18} & \multicolumn{1}{c}{62.53} \\ \hline\hline
MT~\cite{tarvainen2017mean}                       & \multicolumn{1}{c|}{20\%} & \multicolumn{1}{c|}{80\%} & $\times$    & \multicolumn{1}{c|}{73.17} & \multicolumn{1}{c|}{80.29} & \multicolumn{1}{c|}{82.33} & \multicolumn{1}{c}{56.88} \\ 
CPS~\cite{chen2021semi}                      & \multicolumn{1}{c|}{20\%} & \multicolumn{1}{c|}{80\%} & $\times$    & \multicolumn{1}{c|}{73.62} & \multicolumn{1}{c|}{82.98} & \multicolumn{1}{c|}{81.32} & \multicolumn{1}{c}{56.57} \\ 
ICT~\cite{verma2022interpolation}                      & \multicolumn{1}{c|}{20\%} & \multicolumn{1}{c|}{80\%} & $\times$    & \multicolumn{1}{c|}{73.26} & \multicolumn{1}{c|}{76.52} & \multicolumn{1}{c|}{81.03} & \multicolumn{1}{c}{62.22} \\ \hline\hline
\textbf{SGPA-Net (Ours)}                     & \multicolumn{1}{c|}{\textbf{20\%}} & \multicolumn{1}{c|}{\textbf{80\%}} & $\checkmark$    & \multicolumn{1}{c|}{\textbf{80.86}} & \multicolumn{1}{c|}{\textbf{88.57}} & \multicolumn{1}{c|}{\textbf{85.87}} & \multicolumn{1}{c}{\textbf{68.12}} \\ \hline\hline
FD-Sup                   & \multicolumn{1}{c|}{30\%} & \multicolumn{1}{c|}{0\%} & $\times$    & \multicolumn{1}{c|}{73.97} & \multicolumn{1}{c|}{84.12} & \multicolumn{1}{c|}{78.17} & \multicolumn{1}{c}{59.63} \\ \hline\hline
MT~\cite{tarvainen2017mean}                       & \multicolumn{1}{c|}{30\%} & \multicolumn{1}{c|}{70\%} & $\times$    & \multicolumn{1}{c|}{75.92} & \multicolumn{1}{c|}{79.09} & \multicolumn{1}{c|}{84.19} & \multicolumn{1}{c}{64.46} \\
CPS~\cite{chen2021semi}                      & \multicolumn{1}{c|}{30\%} & \multicolumn{1}{c|}{70\%} & $\times$    & \multicolumn{1}{c|}{74.56} & \multicolumn{1}{c|}{82.48} & \multicolumn{1}{c|}{81.50} & \multicolumn{1}{c}{59.69} \\
ICT~\cite{verma2022interpolation}                      & \multicolumn{1}{c|}{30\%} & \multicolumn{1}{c|}{70\%} & $\times$    & \multicolumn{1}{c|}{74.75} & \multicolumn{1}{c|}{83.50} & \multicolumn{1}{c|}{76.01} & \multicolumn{1}{c}{64.74}\\ \hline \hline
\textbf{SGPA-Net (Ours)}                     & \multicolumn{1}{c|}{\textbf{30\%}} & \multicolumn{1}{c|}{\textbf{70\%}} & $\checkmark$    & \multicolumn{1}{c|}{\textbf{81.24}} & \multicolumn{1}{c|}{\textbf{87.91}} & \multicolumn{1}{c|}{\textbf{86.66}} & \multicolumn{1}{c}{\textbf{69.16}} \\ \hline \bottomrule[1pt]
\end{tabular}}
\end{table*}
\begin{figure}[h]
    \centerline{\includegraphics[width=0.5\textwidth,height=0.5\textwidth]{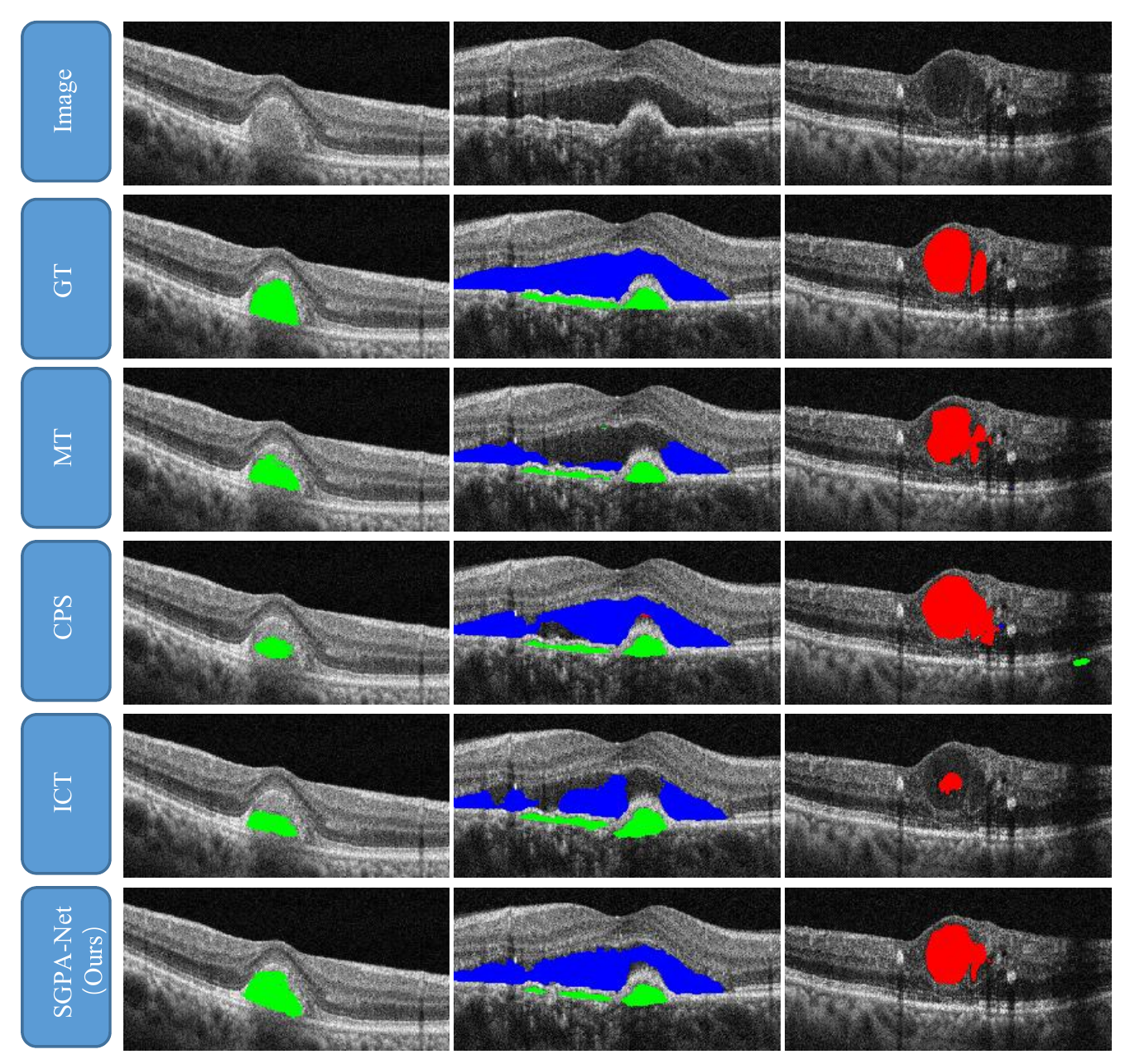}}
\caption{Visualized segmentation results of different semi-supervised methods under 30$\%$ FD setting and 70$\%$ PD setting on the private dataset. From top to bottom are Image, GT, MT, CPS, ICT, SGPA-Net (Ours).}
\label{fig3}
\end{figure}

In order to identify label noise, we adopt the prune by class noise rate (PBNR) strategy, which works by removing examples with a high probability of being mislabeled for every non-diagonal in the $\mathbf{Q}_{\tilde y,y^*}[i][j]$ and select $n \cdot \mathbf{Q}_{\tilde y,y^*}[i][j]$  as mislabeled pixels. Considering our task is multi-category segmentation, we sort the returned error labels index by self-confidence (predicted probability of the given label) for each pixel and select the top 80\% of error labels and error labels with trust values less than 0.8 in the label trust map to form the binary estimate error map $\mathbf{X}_{err}$, where "1" denotes that this pixel is identified as a mislabeled one. Such pixel-level error map $\mathbf{X}_{err}$ can guide the subsequent label and trust map refinement processes.

\subsubsection{Label Refinement and Label Trust Map Refinement}
We highly trust the accuracy of the estimated error map $\mathbf{X}_{err}$ and impose the hard refinement on the given pseudo-labels $\mathbf{\tilde{Y}}$. The predicted label $\mathbf{\hat{Y}} = \{\hat{Y}\}_{i=1}^{n}$ is calculated by the prediction probability $\mathbf{\hat{P}}$:
\begin{equation}
    \mathbf{\hat{Y}}_i=\mathop{\arg\max}\limits_{c} \mathbf{\hat{P}}(c, n).
\end{equation}

We denote $\mathbf{\Dot{Y}} = \{\Dot{Y}\}_{i=1}^{n}$ as the refined-label, which is formulated by:
\begin{equation}
    \dot{Y}_i = \mathbbm{I}(\mathbf{X}_{err}^i = 0)\tilde{Y}_i + \mathbbm{I}(\mathbf{X}_{err}^i = 1)\hat{Y}_i.
\end{equation}

Similar to the pseudo-label $\mathbf{\Tilde{Y}}$, the label trust map $\mathbf{U}$ requires modification since the previous map represented the trustworthiness of the unrefined label. We denote $\mathbf{\Dot{U}}=\{\Dot{U}\}_{i=1}^{n}$ as the refined label trust map, it can be formulated as:
\begin{equation}
    \dot{U}_i=\mathbbm{I}(\mathbf{X}_{err}^i = 0)U_i + \mathbbm{I}(\mathbf{X}_{err}^i = 1)\delta,
\end{equation}
Where $\delta \in [0, 1]$ represents the confidence level for the estimated error map $\mathbf{X}_{err}$, and we set it to 1.

\subsection{Training SCLGPA-Net Based on Mean-Teacher Architecture}

We incorporate the updated information from $\{\mathbf{\Dot{Y}}, \mathbf{\Dot{U}}\}$ into the point-annotated data ($\{ \mathbf{\Tilde{Y}},\mathbf{U}\}$ of PD) as part of our SCLGPA-Net training process. It's important to note that the experimental parameters utilized during the SGPA-Net training phase were retained and applied in the training of the SCLGPA-Net as well. This continuity in parameter settings ensures consistency and effectiveness throughout the model's development.

\begin{figure*}[!t]
\hspace{5mm}
 \begin{minipage}{0.23\linewidth}
 	\vspace{3pt}
 	\centerline{\includegraphics[width=\textwidth]{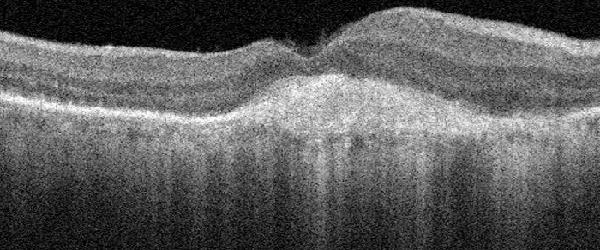}}
 	\vspace{3pt}
 	\centerline{\includegraphics[width=\textwidth]{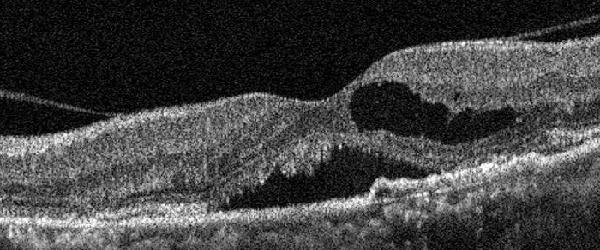}}
 	\vspace{3pt}
  \centerline{(a)}
 \end{minipage}
\begin{minipage}{0.23\linewidth}
 	\vspace{3pt}
 	\centerline{\includegraphics[width=\textwidth]{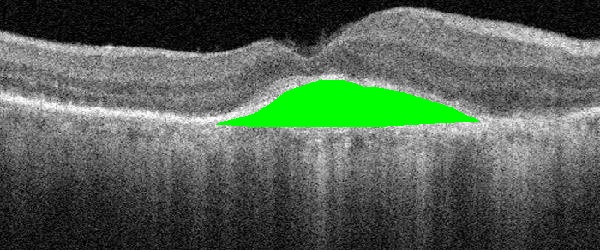}}
 	\vspace{3pt}
 	\centerline{\includegraphics[width=\textwidth]{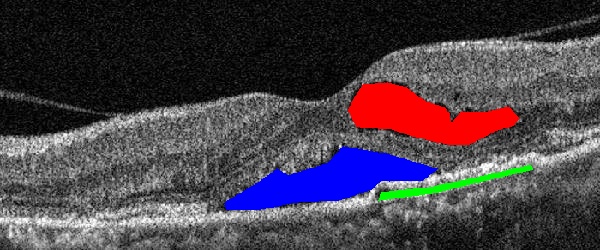}}
 	\vspace{3pt}
  \centerline{(b)}
  
 \end{minipage}
\begin{minipage}{0.23\linewidth}
 	\vspace{3pt}
 	\centerline{\includegraphics[width=\textwidth]{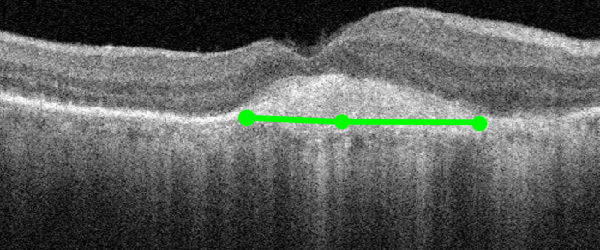}}
 	\vspace{3pt}
 	\centerline{\includegraphics[width=\textwidth]{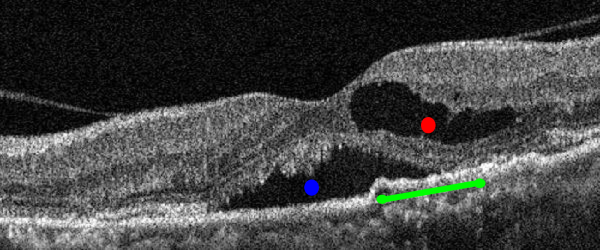}}
 	\vspace{3pt}
  \centerline{(c)}
  
 \end{minipage}
\begin{minipage}{0.23\linewidth}
 	\vspace{3pt}
 	\centerline{\includegraphics[width=\textwidth]{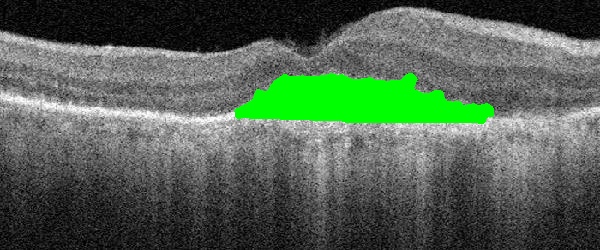}}
 	\vspace{3pt}
 	\centerline{\includegraphics[width=\textwidth]{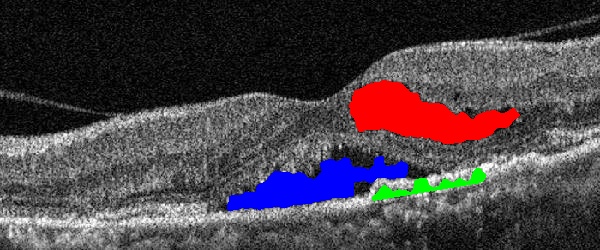}}
 	\vspace{3pt}
  \centerline{(d)}
  
 \end{minipage}
 \caption{Examples of retinal in OCT image with manual annotations on the private dataset. (a) The original OCT image with fluids; (b) The full annotations. (c) The point annotations; (d) The pseudo-labels generated by SGPLG. In (b) and (d), the green, blue, and red contours denote PED, SRF, and IRF, respectively.}
\label{fig4}
 \end{figure*}

\begin{table*}[!t]
\centering
    \renewcommand\arraystretch{1.5}
    \caption{Oct Fluid Segmentation Studies on Private Dataset. Comparison of the Experimental Results of Other Label-Denoising Methods on Different Ratios of FD and PD. The Best Results are in Bold. (Dice Unit: $\%$)}
    \label{tab2}
    \resizebox{0.8\linewidth}{!}{
\begin{tabular}{c|lll|llllll}
\toprule[1pt]
\hline
\multirow{2}{*}{\textbf{Methods}} & \multicolumn{3}{c|}{\textbf{Settings}}                                                       & \multicolumn{4}{c}{\textbf{Metrics}}                                                                                                                                                                                    \\ \cline{2-8} 
\multicolumn{1}{c|}{}                         & \multicolumn{1}{c|}{FD}    & \multicolumn{1}{c|}{PD}    & Separate        & \multicolumn{1}{c|}{DSC}      & \multicolumn{1}{c|}{D$_{SRF}$}      & \multicolumn{1}{c|}{D$_{IRF}$}      & \multicolumn{1}{c}{D$_{PED}$} \\ \hline \hline
FD\&PD-Sup                                    & \multicolumn{1}{c|}{20$\%$}  & \multicolumn{1}{c|}{80$\%$}  & \multicolumn{1}{c|}{$\times$}     & \multicolumn{1}{c|}{77.84}          & \multicolumn{1}{c|}{84.28}          & \multicolumn{1}{c|}{85.94}          & \multicolumn{1}{c}{63.31} \\ \hline
Co-teaching~\cite{han2018co}                                  & \multicolumn{1}{c|}{20$\%$}  & \multicolumn{1}{c|}{80$\%$}  & \multicolumn{1}{c|}{$\times$}     & \multicolumn{1}{c|}{80.07}          & \multicolumn{1}{c|}{87.79} & \multicolumn{1}{c|}{83.51}          & \multicolumn{1}{c}{68.91}\\
TriNet\cite{zhang2020robust}                                        & \multicolumn{1}{c|}{20$\%$}  & \multicolumn{1}{c|}{80$\%$}  & \multicolumn{1}{c|}{$\times$}     & \multicolumn{1}{c|}{80.00}          & \multicolumn{1}{c|}{\textbf{88.02}}          & \multicolumn{1}{c|}{83.13}          & \multicolumn{1}{c}{68.86} \\ 
2SRnT\cite{zhang2020characterizing}                                         & \multicolumn{1}{c|}{20$\%$}  & \multicolumn{1}{c|}{80$\%$}  & \multicolumn{1}{c|}{$\times$}     & \multicolumn{1}{c|}{78.97}          & \multicolumn{1}{c|}{84.37}          & \multicolumn{1}{c|}{85.92}          & \multicolumn{1}{c}{66.63}         \\ \hline
MTCL~\cite{xu2022anti}                                         & \multicolumn{1}{c|}{20$\%$}  & \multicolumn{1}{c|}{80$\%$}  & \multicolumn{1}{c|}{$\checkmark$} & \multicolumn{1}{c|}{78.67}          & \multicolumn{1}{c|}{83.71}          & \multicolumn{1}{c|}{87.43} & \multicolumn{1}{c}{64.88} \\
Dast\cite{yang2022learning}                                          & \multicolumn{1}{c|}{20$\%$}  & \multicolumn{1}{c|}{80$\%$}  & \multicolumn{1}{c|}{$\checkmark$} & \multicolumn{1}{c|}{75.41}          & \multicolumn{1}{c|}{79.09}          & \multicolumn{1}{c|}{77.70}          & \multicolumn{1}{c}{69.45}            \\ \hline \hline
\textbf{SCLGPA-Net (Ours)}                                       & \multicolumn{1}{c|}{\textbf{20$\%$}}  & \multicolumn{1}{c|}{\textbf{80$\%$}}  & \multicolumn{1}{c|}{$\checkmark$} & \multicolumn{1}{c|}{\textbf{82.07}} & \multicolumn{1}{c|}{86.98}          & \multicolumn{1}{c|}{\textbf{89.03}}          & \multicolumn{1}{c}{\textbf{70.28}}\\ \hline \hline

FD\&PD-Sup                                    & \multicolumn{1}{c|}{30$\%$}  & \multicolumn{1}{c|}{70$\%$}  & \multicolumn{1}{c|}{$\times$}     & \multicolumn{1}{c|}{79.32}          & \multicolumn{1}{c|}{87.08}          & \multicolumn{1}{c|}{83.75}          & \multicolumn{1}{c}{67.13} \\ \hline \hline
Co-teaching~\cite{han2018co}                                  & \multicolumn{1}{c|}{30$\%$}  & \multicolumn{1}{c|}{70$\%$}  & \multicolumn{1}{c|}{$\times$}     & \multicolumn{1}{c|}{80.66}          & \multicolumn{1}{c|}{86.92}          & \multicolumn{1}{c|}{86.63} & \multicolumn{1}{c}{68.43}\\
TriNet\cite{zhang2020robust}                                        & \multicolumn{1}{c|}{30$\%$}  & \multicolumn{1}{c|}{70$\%$}  & \multicolumn{1}{c|}{$\times$}     & \multicolumn{1}{c|}{80.00} & \multicolumn{1}{c|}{85.41}          & \multicolumn{1}{c|}{83.03}          & \multicolumn{1}{c}{\textbf{71.54}}        \\ 
2SRnT\cite{zhang2020characterizing}                                         & \multicolumn{1}{c|}{30$\%$}  & \multicolumn{1}{c|}{70$\%$}  & \multicolumn{1}{c|}{$\times$}     & \multicolumn{1}{c|}{79.16}          & \multicolumn{1}{c|}{88.39}          & \multicolumn{1}{c|}{86.67}          & \multicolumn{1}{c}{62.42}          \\ \hline
MTCL~\cite{xu2022anti}                                         & \multicolumn{1}{c|}{30$\%$}  & \multicolumn{1}{c|}{70$\%$}  & \multicolumn{1}{c|}{$\checkmark$} & \multicolumn{1}{c|}{80.35}          & \multicolumn{1}{c|}{88.19}          & \multicolumn{1}{c|}{83.25}          & \multicolumn{1}{c}{69.59} \\
Dast\cite{yang2022learning}                                          & \multicolumn{1}{c|}{30$\%$}  & \multicolumn{1}{c|}{70$\%$}  & \multicolumn{1}{c|}{$\checkmark$} & \multicolumn{1}{c|}{76.83}          & \multicolumn{1}{c|}{80.05}          & \multicolumn{1}{c|}{85.01}          & \multicolumn{1}{c}{65.43} \\ \hline \hline
\textbf{SCLGPA-Net (Ours)}                                      & \multicolumn{1}{c|}{\textbf{30$\%$}}  & \multicolumn{1}{c|}{\textbf{70$\%$}}  & \multicolumn{1}{c|}{$\checkmark$} & \multicolumn{1}{c|}{\textbf{82.87}}          & \multicolumn{1}{c|}{\textbf{89.46}} & \multicolumn{1}{c|}{\textbf{88.78}}          & \multicolumn{1}{c}{70.37}\\ \hline \hline
PD-Sup                                        & \multicolumn{1}{c|}{0$\%$}   & \multicolumn{1}{c|}{100$\%$} & \multicolumn{1}{c|}{$\times$}                         & \multicolumn{1}{c|}{73.64}          & \multicolumn{1}{c|}{76.67}          & \multicolumn{1}{c|}{85.03}         & \multicolumn{1}{c}{59.22} \\
FD-Sup                                        & \multicolumn{1}{c|}{100$\%$} & \multicolumn{1}{c|}{0$\%$}   & \multicolumn{1}{c|}{$\times$}                         & \multicolumn{1}{c|}{83.53}          & \multicolumn{1}{c|}{89.40}          & \multicolumn{1}{c|}{87.88}          & \multicolumn{1}{c}{73.31}   \\ \hline \bottomrule[1pt]
\end{tabular}}
\end{table*}
\begin{figure*}[!t]
    \centerline{\includegraphics[width=\textwidth,height=0.9\textwidth]{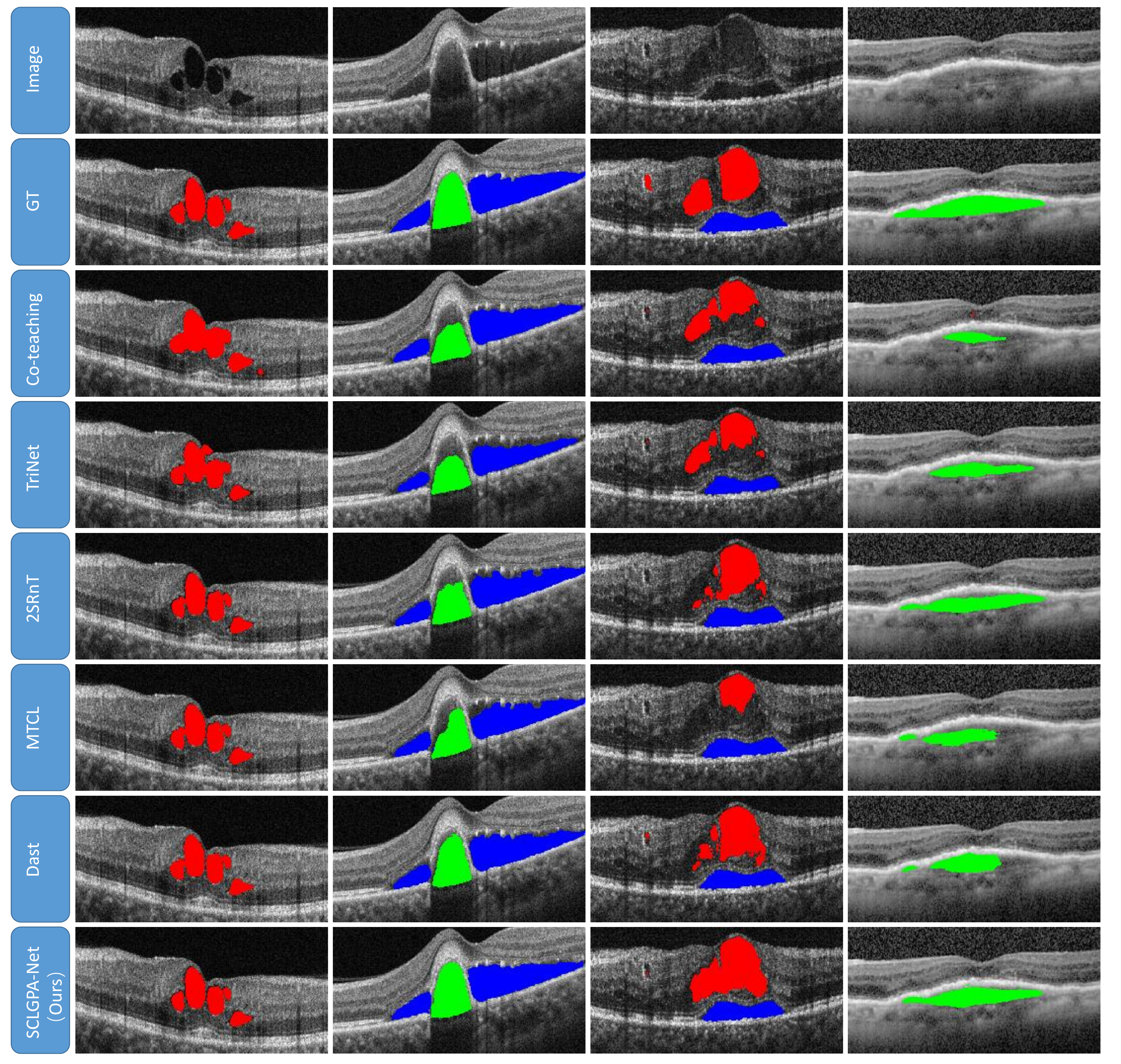}}
\caption{Visualized segmentation results of different label-denoising methods under 30$\%$ FD setting and 70$\%$ PD setting on the private dataset. From top to bottom are Image, GT, Co-teaching, TriNet, 2SRnT, MTCL, Dast, and SCLGPA-Net (Ours).}

\label{fig5}
\end{figure*}

\subsection{Final Loss Function}
The loss function for training SGPA-Net and SCLGPA-Net as a whole remains consistent. In general, the total loss is divided into three parts: the supervised loss $\mathcal{L}_f = \mathcal{L}_f^{ce} + \mathcal{L}_f^{dice}$ on FD, the perturbation-based consistency loss $\mathcal{L}_{con}$ and the supervised loss $\mathcal{L}_p = \mathcal{L}_p^{ce} \cdot \mathbf{U}' + \mathcal{L}_p^{dice}$ on PD. The total loss is calculated by:
\begin{equation}
    \mathcal{L} = \alpha \mathcal{L}_f + \beta \mathcal{L}_p + \lambda \mathcal{L}_{con}.
    \label{loss}
\end{equation}

Here, $\mathbf{U}^{'}$ in $\mathcal{L}_p$ is $\mathbf{U}$ for training SGPA-Net and will be replaced by $\mathbf{\dot{U}}$ for SCLGPA-Net. Empirically, $\alpha$ and $\beta$ are hyper-parameters and we set $\alpha = 1$, $\beta = 1$. The $\mathcal{L}_{con}$ are calculated by the pixel-wise mean squared error (MSE), $\lambda$ is a ramp-up trade-off weight commonly scheduled by the time-dependent Gaussian function~\cite{cui2019semi} $\lambda(t) = w_{max} \cdot e^{(-5(1-\frac{t}{t_{max}})^2)}$, where $w_{max}$ is the maximum weight commonly set as 0.1~\cite{yu2019uncertainty} and $t_{max}$ is the maximum training iteration. Such a $\lambda$ weight representation avoids being dominated by misleading targets when starting online training.

\section{Experiments}
\subsection{Private Dataset and Experimental Setup}

Our data come from the Eye Center at the Second Affiliated Hospital, School of Medicine, Zhejiang University. The dataset consists of OCT images from various patients, taken at different times and with two distinct resolutions of 1476 $\times$ 560 and 1520 $\times$ 596. Given the extensive background area in the original images, any other fluids appeared comparatively small. To address this, we centrally cropped all images and resized them to a resolution of 600 $\times$ 250. A subset of OCT images rich in the fluid was selected for comprehensive and weak point labeling. The total dataset consists of 1704 OCT images, with 1304 images designated for training and 400 for testing. To ensure the reliability of our results, we meticulously partitioned the dataset such that data from a single patient was used exclusively for either training or testing. 

\begin{table*}[!t]
\centering
\small
    \renewcommand\arraystretch{1.5}
    \caption{Oct Fluid Segmentation Studies on RETOUCH Dataset. Comparison of the Experimental Results of Other Semi-Supervised and Label-Denoising Methods. The Best Results are in Bold. (Dice Unit: $\%$)}
    \label{tab3}
    \large
\scalebox{0.8}{
\begin{tabular}{cc|cc|c|cccc}
\toprule[1pt]
\hline
\multicolumn{2}{c|}{\multirow{2}{*}{\textbf{Methods}}}                        & \multicolumn{2}{c|}{\textbf{Spectralis}}                & \multirow{2}{*}{\textbf{Point-Annotation}} & \multicolumn{4}{c}{\textbf{Metrics}}                                                                                                       \\ \cline{3-4} \cline{6-9} 
\multicolumn{2}{c|}{}                                                & \multicolumn{1}{c|}{FD}    & \multicolumn{1}{c|}{PD}    &         & \multicolumn{1}{c|}{DSC}      & \multicolumn{1}{c|}{D$_{SRF}$}      & \multicolumn{1}{c|}{D$_{IRF}$}      & \multicolumn{1}{c}{D$_{PED}$}           \\ \hline \hline
\multicolumn{2}{c|}{FD-Sup}                                          & \multicolumn{1}{c|}{10}      & 0         & $\times$                      & \multicolumn{1}{c|}{76.92}          & \multicolumn{1}{c|}{83.04}          & \multicolumn{1}{c|}{66.92}          & 80.80          \\ \hline \hline
\multicolumn{1}{c|}{\multirow{4}{*}{\textbf{Semi-Supervised}}}  & MT~\cite{tarvainen2017mean}          & \multicolumn{1}{c|}{10}      & 20        & $\times$                     & \multicolumn{1}{c|}{77.20}          & \multicolumn{1}{c|}{83.11}          & \multicolumn{1}{c|}{67.91}          & 80.57          \\ 
\multicolumn{1}{c|}{}                                  & CPS~\cite{chen2021semi}          & \multicolumn{1}{c|}{10}      & 20        & $\times$                     & \multicolumn{1}{c|}{77.02}          & \multicolumn{1}{c|}{83.29}          & \multicolumn{1}{c|}{67.04}          & 80.72          \\ 
\multicolumn{1}{c|}{}                                  & ICT~\cite{verma2022interpolation}          & \multicolumn{1}{c|}{10}      & 20        & $\times$                     & \multicolumn{1}{c|}{77.07}               & \multicolumn{1}{c|}{\textbf{85.36}}               & \multicolumn{1}{c|}{66.38} & \multicolumn{1}{c}{79.47} \\
\multicolumn{1}{c|}{}                                  & \textbf{SGPA-Net (Ours)}    & \multicolumn{1}{c|}{\textbf{10}}      & \textbf{20}        & $\checkmark$                     & \multicolumn{1}{c|}{\textbf{79.39}} & \multicolumn{1}{c|}{85.29} & \multicolumn{1}{c|}{\textbf{71.13}} & \textbf{81.75} \\ \hline \hline
\multicolumn{2}{c|}{FD \& PD-Sup}                                   & \multicolumn{1}{c|}{10}      & 20        & $\checkmark$                     & \multicolumn{1}{c|}{78.02}          & \multicolumn{1}{c|}{85.09}          & \multicolumn{1}{c|}{69.48}          & 79.49          \\ \hline \hline
\multicolumn{1}{c|}{\multirow{6}{*}{\textbf{Label-Denoising}}} & Co-teaching~\cite{han2018co} & \multicolumn{1}{c|}{10}      & 20        & $\checkmark$                     & \multicolumn{1}{c|}{79.22}          & \multicolumn{1}{c|}{85.74}          & \multicolumn{1}{c|}{70.38}          & 81.53          \\ 
\multicolumn{1}{c|}{}                                  & TriNet\cite{zhang2020robust}       & \multicolumn{1}{c|}{10}      & 20        & $\checkmark$                     & \multicolumn{1}{c|}{78.74}          & \multicolumn{1}{c|}{87.59}          & \multicolumn{1}{c|}{68.71}          & 79.91          \\ 
\multicolumn{1}{c|}{}                                  & 2SRnT\cite{zhang2020characterizing}       & \multicolumn{1}{c|}{10}      & 20        & $\checkmark$                     & \multicolumn{1}{c|}{79.49}          & \multicolumn{1}{c|}{84.35}          & \multicolumn{1}{c|}{71.49}          & 82.64          \\ 
\multicolumn{1}{c|}{}                                  & MTCL~\cite{xu2022anti}         & \multicolumn{1}{c|}{10}      & 20        & $\checkmark$                     & \multicolumn{1}{c|}{79.83}          & \multicolumn{1}{c|}{\textbf{87.83}} & \multicolumn{1}{c|}{71.61}          & 80.03          \\ 
\multicolumn{1}{c|}{}                                  & Dast\cite{yang2022learning}        & \multicolumn{1}{c|}{10}      & 20        & $\checkmark$                     & \multicolumn{1}{c|}{78.98}          & \multicolumn{1}{c|}{84.96}          & \multicolumn{1}{c|}{70.94}          & 81.03          \\ 
\multicolumn{1}{c|}{}                                  & \textbf{SCLGPA-Net (Ours)}  & \multicolumn{1}{c|}{\textbf{10}}      & \textbf{20}        & $\checkmark$                     & \multicolumn{1}{c|}{\textbf{80.89}} & \multicolumn{1}{c|}{85.43}          & \multicolumn{1}{c|}{\textbf{73.16}} & \textbf{84.08} \\ \hline \hline
\multicolumn{2}{c|}{FD-Sup}                                   & \multicolumn{1}{c|}{30}      & 0        & $\times$                     & \multicolumn{1}{c|}{81.63}          & \multicolumn{1}{c|}{89.66}          & \multicolumn{1}{c|}{73.92}          & 81.31          \\ \hline
\bottomrule[1pt]
\end{tabular}}
\end{table*}


\subsubsection{Baseline Approaches}
We consider three different baselines to fully compare the effectiveness of our method. The baselines can be categorized as follows:
\begin{itemize}
\item[$\bullet$] Fully-Supervised baselines: $\textbf{FD-Sup}$: uses only FD to train the backbone (2D U-Net) network; $\textbf{PD-Sup}$: uses only PD to train the backbone network;
$\textbf{FD\&PD-Sup}$: mixes both FD and PD to train the backbone network.
\end{itemize}

\begin{itemize}
\item[$\bullet$] SSL baselines: $\textbf{MT}$~\cite{tarvainen2017mean}: encourages prediction consistency between the student model and the teacher model; $\textbf{CPS}$~\cite{chen2021semi}: uses two networks with the same structure but different initialization, adding constraints to ensure that the output of both networks for the same sample exhibits similarity; $\textbf{ICT}$~\cite{verma2022interpolation}: encourages the coherence between the prediction at an interpolation of unlabeled points and the interpolation of the predictions at those points.
\end{itemize}

\begin{itemize}
\item[$\bullet$] Label-Denoising baselines: $\textbf{2SRnT}$\cite{zhang2020characterizing}: involves two stages for pre-training a network using a combination of different datasets, followed by fine-tuning the labels using confidence estimates to train a second network; $\textbf{Co-teaching}$\cite{han2018co}: a joint teaching method of the double network; $\textbf{TriNet}$\cite{zhang2020robust}: a tri-network based noise-tolerant method extended from Co-teaching. Separate FD and PD: $\textbf{MTCL}$\cite{xu2022anti}: Mean-Teacher-assisted Confident Learning, which can robustly learn segmentation from limited high-quality labeled data and abundant low-quality labeled data; $\textbf{Dast}$\cite{yang2022learning}: a dual-branch network to separately learn from the accurate and noisy labels.
\end{itemize}
\begin{figure}[!t]
\centerline{\includegraphics[width=\linewidth]{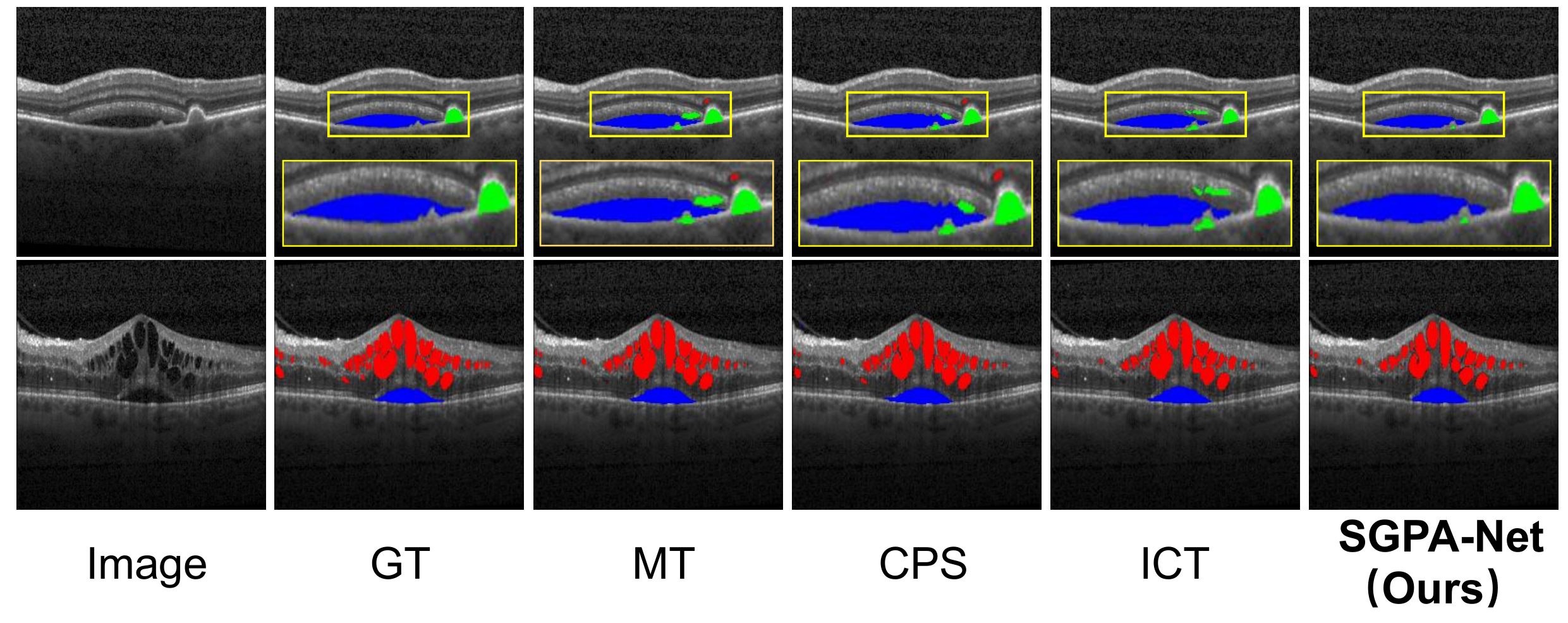}}
    \caption{Visualized segmentation results of different semi-supervised methods on RETOUCH dataset. From left to right are Image, GT, MT, CPS, ICT, SGPA-Net (Ours).}
    \label{fig6}
\end{figure}
The effectiveness of the SGPLG module in our approach was validated through comparisons with SSL methods. Similarly, the effectiveness of our CLGLR module was validated through comparisons with Label-Denoising methods.

\subsubsection{Implementation and Evaluation Metric}

\begin{figure*}[!t]
\centerline{\includegraphics[width=\linewidth]{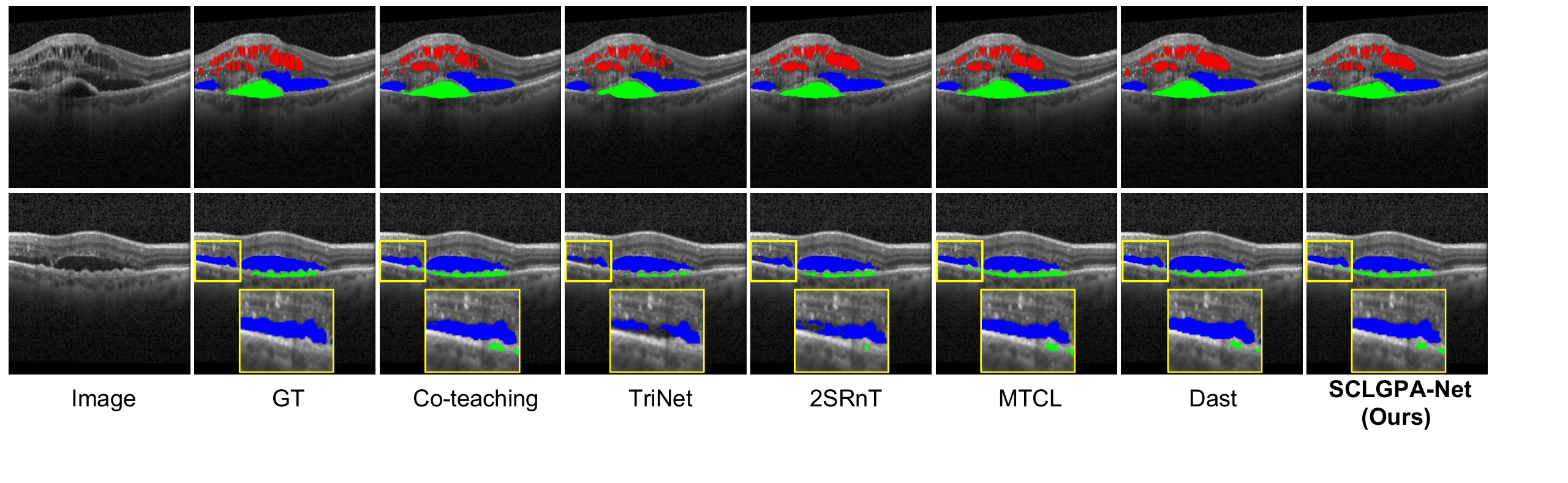}}
    \caption{Visualized segmentation results of different label-denoising methods on RETOUCH dataset. From left to right are Image, GT, Co-teaching, TriNet, 2SRnT, MTCL, Dast, and SCLGPA-Net (Ours).}
    \label{fig7}
\end{figure*}

\subsection{Experiments on Private Dataset}
To ensure the reliability of our experiments and verify the gain of our different modules on network performance, we employ two different validation approaches. The first approach involves evaluating our method in a semi-supervised setting, where we solely consider the pseudo-labels generated by the SGPLG module as additional knowledge and compare it with other semi-supervised methods. The second approach incorporates the pseudo-labels as prior knowledge, integrating them into the CLGLR module, and comparing it with other label-denoising methods.

\subsubsection{Compared With Other Semi-Supervised Methods}
Table~\ref{tab1} presents the results of an extensive study on semi-supervised OCT fluid segmentation methods, assessing their performance under various data settings. When evaluating metrics such as DSC, D$_{SRF}$, D$_{IRF}$, D$_{PED}$, it becomes evident that SGPA-Net's performance significantly surpasses that of other semi-supervised methods. Our method stands out for its ability to exploit point annotation information from unlabeled data. This capability puts our approach significantly ahead of other semi-supervised methods. It reaffirms the efficacy of our proposed SGPLG module in utilizing point annotations to generate reliable pseudo-labels for unlabeled data. 
Fig~\ref{fig3} shows the visualization results of all methods. These visualizations clearly demonstrate that SGPA-Net delivers superior segmentation outcomes. Furthermore, Fig~\ref{fig4} shows the visualization of pseudo-labels generated from point annotations, we can observe that the results are acceptable.

 \subsubsection{Compared With Other Label-Denoising Methods }
We consider the generated pseudo-labels (excluding the label trust map) as prior knowledge for comparison with other label-denoising methods.
Table~\ref{tab2} presents the comparison results under 20$\%$, and 30$\%$ FD settings. Firstly, in the typical supervised settings of FD-Sup and FD\&PD-Sup, the network performs poorly and can benefit from additional PD, although their labels contain misinformation. We hypothesize two possibilities:
(i) The partially pseudo-labels generated by SGPLG are highly accurate and can provide reliable guidance for the network. The effect of PD-Sup can also reach 73.64\%, which is also confirmed.
(ii) Even with only 30$\%$ of the FD data, the network may still be under-fitting and potentially learn valuable features from the PD.

When turning to the Mix FD and PD setting, the three baseline methods, 2SRnT, Co-teaching, and TriNet have shown effective performance in mitigating the negative effects caused by pseudo-labels. In contrast, under the Separate FD and PD settings, MTCL has demonstrated a steady improvement ranging from 20$\%$ to 30$\%$. Although Dast has also shown improvement, it still falls short of the baseline performance (FD\&PD-Sup). One possible explanation for this discrepancy is domain crossing since Dast was originally designed to perform COVID-19 pneumonia lesion segmentation. Under the 20$\%$ FD setting, Co-teaching and MTCL achieved highly competitive results, but SCLGPA-Net still outperforms them in most metrics. When we increased the proportion of FD to 30$\%$, our method substantially surpassed other label-denoising methods. Overall, in the OCT fluid segmentation task, our method achieves satisfactory results, suggesting that the label trust map and estimated error map can accurately characterize the location of incorrect labels, enabling the network to fully exploit these informative refined-labels. Fig.~\ref{fig5} presents the results of SCLGPA-Net and other approaches under the 30$\%$ FD setting and 70$\%$ PD setting. It is evident that the mask predicted by our method is closer to the ground truth, further demonstrating the effectiveness of SCLGPA-Net. 

\begin{table*}[!t]
\centering
    \renewcommand\arraystretch{1.5}
    \caption{Performance Comparison on RETOUCH with Other Fluid Segmentation Methods. The Best Results are in Bold. (mIoU ± standard deviation)}
    \label{tab4}
    \large
\scalebox{0.8}{
\begin{tabular}{c|cccc}
\toprule[1pt]
\hline
\textbf{Methods}                    & \multicolumn{1}{c|}{$N_l$=3}      & \multicolumn{1}{c|}{$N_l$=6}      & \multicolumn{1}{c|}{$N_l$=12}     & $N_l$=24     \\ \hline \hline
Baseline                   & \multicolumn{1}{c|}{0.15 ± 0.07} & \multicolumn{1}{c|}{0.27 ± 0.08} & \multicolumn{1}{c|}{0.35 ± 0.06} & 0.49 ± 0.05 \\ \hline \hline
IIC (Ji, Henriques, and Vedaldi 2019)~\cite{ji2019invariant}                        & \multicolumn{1}{c|}{0.22 ± 0.09} & \multicolumn{1}{c|}{0.32 ± 0.07} & \multicolumn{1}{c|}{0.41 ± 0.07} & 0.53 ± 0.06 \\ 
Perone and Cohen-Add (2018)~\cite{perone2019unsupervised}
& \multicolumn{1}{c|}{0.21 ± 0.09} & \multicolumn{1}{c|}{0.32 ± 0.07} & \multicolumn{1}{c|}{0.39 ± 0.07} & 0.50 ± 0.08 \\ 
MLDS (2021)~\cite{reiss2021every}                 & \multicolumn{1}{c|}{0.16 ± 0.15} & \multicolumn{1}{c|}{0.35 ± 0.11} & \multicolumn{1}{c|}{0.54 ± 0.09} & 0.59 ± 0.07 \\ 
RPG (2022)~\cite{seibold2022reference}                        & \multicolumn{1}{c|}{0.21 ± 0.10} & \multicolumn{1}{c|}{0.30 ± 0.08} & \multicolumn{1}{c|}{0.45 ± 0.08} & 0.54 ± 0.08 \\
RPG$^+$ (2022)~\cite{seibold2022reference}                        & \multicolumn{1}{c|}{0.31 ± 0.11} & \multicolumn{1}{c|}{0.45 ± 0.10} & \multicolumn{1}{c|}{0.55 ± 0.08} & 0.59 ± 0.08 \\ \hline \hline
\textbf{SGPA-Net (Ours)}            & \multicolumn{1}{c|}{0.490 ± 0.018}   & \multicolumn{1}{c|}{0.532 ± 0.014}   & \multicolumn{1}{c|}{0.544 ± 0.019}   & 0.592 ± 0.015   \\
\textbf{SCLGPA-Net (Ours)}          & \multicolumn{1}{c|}{\textbf{0.520 ± 0.015}}   & \multicolumn{1}{c|}{\textbf{0.549 ± 0.020}}   & \multicolumn{1}{c|}{\textbf{0.563 ± 0.016}}   & \textbf{0.604 ± 0.017}   \\ \hline \hline
Full Access ($N_l$ = 415)     & \multicolumn{4}{c}{0.62 ± 0.05}                                                                        \\ \hline \bottomrule[1pt]
\end{tabular}}
\end{table*}

\subsection{RETOUCH Dataset and Experimental Setup}
RETOUCH~\cite{bogunovic2019retouch} is a publicly available data set for retinal fluid segmentation. The dataset comprises OCT volumes, which are collections of B-scans, showcasing various retinal conditions. These volumes are acquired using imaging devices from three distinct vendors: Spectralis, Cirrus, and Topcon. In general, B-scans exhibit variations in appearance across different manufacturers. 
For our experiments, we selected the image acquired with the Spectralis device as the dataset for our study, which consists of 24 volumes, each containing 49 slices.
We designed two different experimental settings for different comparison methods:

\begin{itemize}
\item[$\bullet$] We randomly selected 30 slices from the first 20 volumes, with 10 slices labeled and 20 slices unlabeled for training. The remaining volumes were used as the test set. We scale the size of the image uniformly to 256 $\times$ 256 and other experimental settings and evaluation metrics are consistent with the private dataset.

\item[$\bullet$] We adopt the methodology outlined in ~\cite{reiss2021every} for the experimental setup. The training set contains 14 OCT volumes, and the remaining 10 volumes serve as the validation set and test set. We perform 10-fold cross-validation with training sets using $N_l$ labeled
images ($N_l \in {3,6,12,24}$) with validation and test sets of equal size. 
Importantly, we ensure that in each split, all different diseases are represented at least once in the mask labels. We scale the size of the image uniformly to 256 $\times$ 256 and the batch size is (2, 4) for labeled and unlabeled images. We use mean Intersection over Union (mIoU) as the performance metric. We evaluate our approach every epoch and the best-performing model on the validation set is applied to the test set.
\end{itemize}

\begin{figure*}[!t]
\hspace{5mm}
 \begin{minipage}{0.23\linewidth}
 	\vspace{3pt}
 	\centerline{\includegraphics[width=\textwidth]{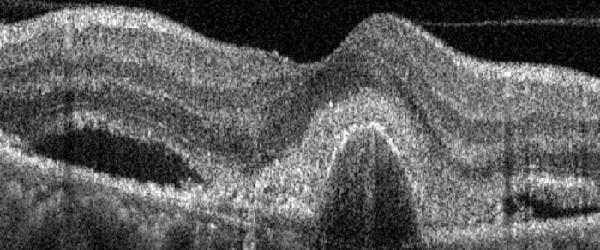}}
 	\vspace{3pt}
 	\centerline{\includegraphics[width=\textwidth]{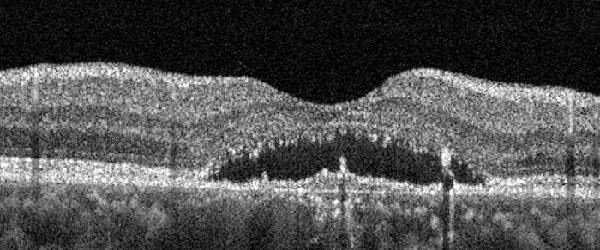}}
 	\vspace{3pt}
 	\centerline{\includegraphics[width=\textwidth]{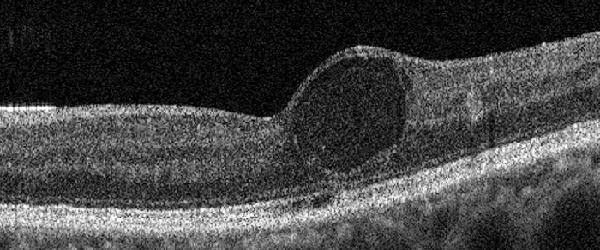}}
 	\vspace{3pt}
 	\centerline{\includegraphics[width=\textwidth]{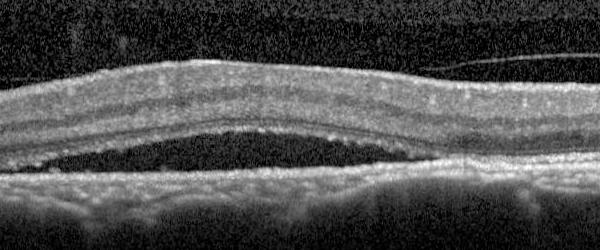}}
 	\centerline{(a)}
 \end{minipage}
 \begin{minipage}{0.23\linewidth}
 	\vspace{3pt}
 	\centerline{\includegraphics[width=\textwidth]{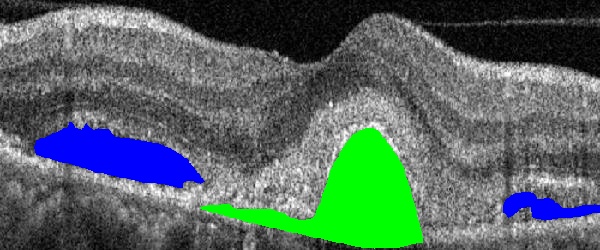}}
 	\vspace{3pt}
 	\centerline{\includegraphics[width=\textwidth]{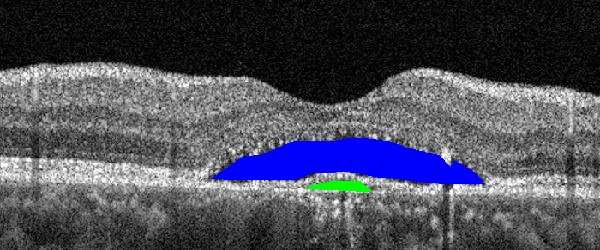}}
 	\vspace{3pt}
 	\centerline{\includegraphics[width=\textwidth]{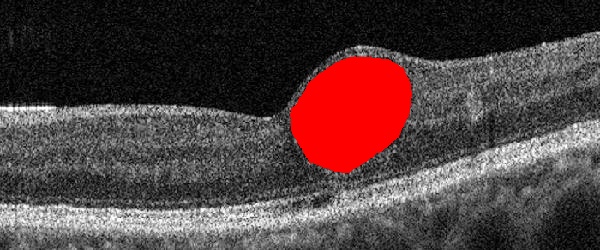}}
 	\vspace{3pt}
 	\centerline{\includegraphics[width=\textwidth]{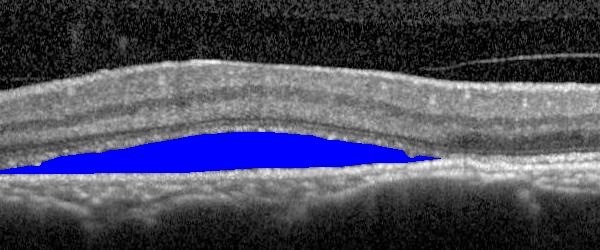}}
 	\centerline{(b)}
 \end{minipage}
  \begin{minipage}{0.23\linewidth}
 	\vspace{3pt}
 	\centerline{\includegraphics[width=\textwidth]{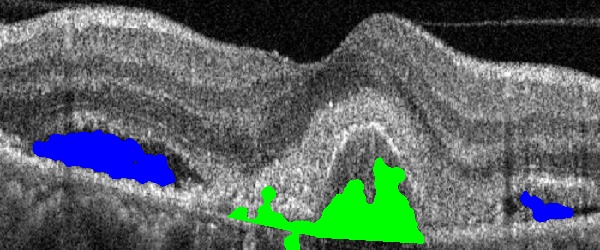}}
 	\vspace{3pt}
 	\centerline{\includegraphics[width=\textwidth]{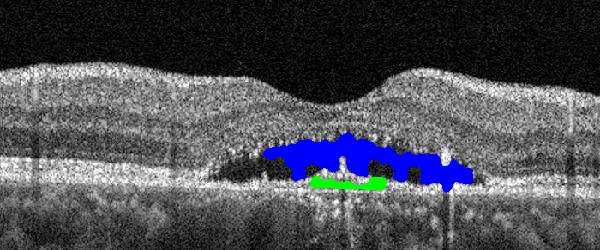}}
 	\vspace{3pt}
 	\centerline{\includegraphics[width=\textwidth]{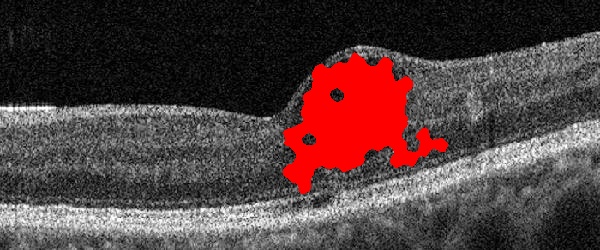}}
 	\vspace{3pt}
 	\centerline{\includegraphics[width=\textwidth]{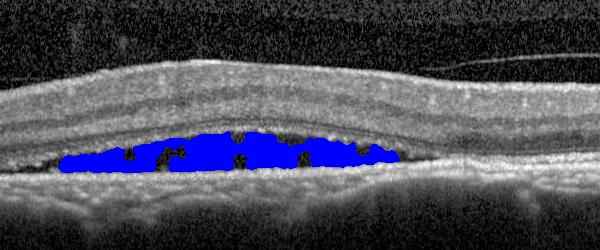}}
 	\centerline{(c)}
 \end{minipage}
   \begin{minipage}{0.23\linewidth}
 	\vspace{3pt}
 	\centerline{\includegraphics[width=\textwidth]{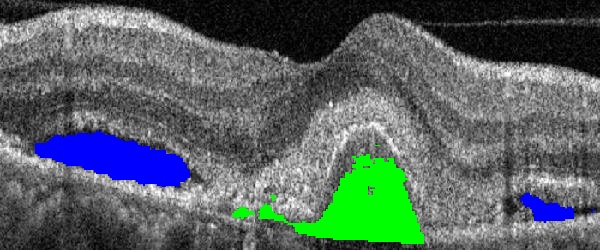}}
 	\vspace{3pt}
 	\centerline{\includegraphics[width=\textwidth]{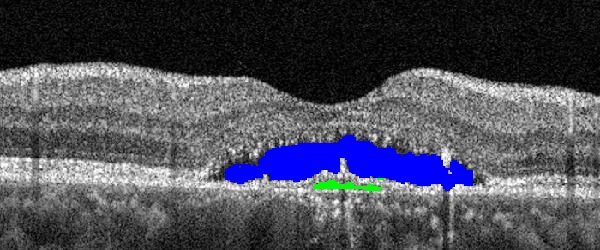}}
 	\vspace{3pt}
 	\centerline{\includegraphics[width=\textwidth]{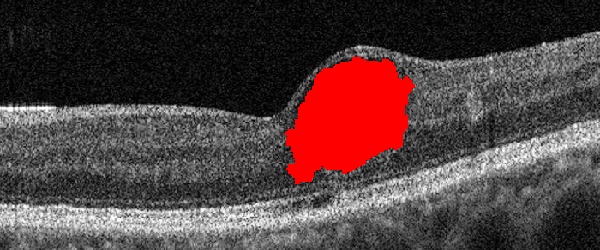}}
 	\vspace{3pt}
 	\centerline{\includegraphics[width=\textwidth]{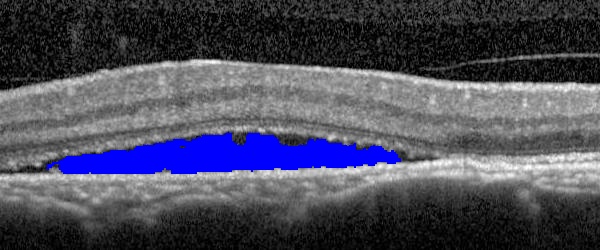}}
 	\centerline{(d)}
 \end{minipage}

\caption{Examples of retinal in OCT image with manual annotations on the private dataset. (a) The original OCT images with fluids; (b) The full-labels; (c) The pseudo-labels generated by SGPLG; (d) The refined-labels by CLGLR.}
\label{fig8}
\end{figure*}

\subsection{Experiments on RETOUCH Dataset}
Similarly to the experiment of the private dataset, we employ two different validation approaches for the first experiment setting. As for the second experiment setting, all experiments in Seibold~\textit{et al.}~\cite{seibold2022reference} as comparative experimental.

\subsubsection{Comparative Experiments with Other Semi-Supervised and Label-Denoising Methods}
Table~\ref{tab3} provides the experimental results of OCT fluid segmentation conducted on the RETOUCH dataset. In the semi-supervised methods, approaches such as MT, CPS, and ICT utilized unlabeled data without the assistance of point annotation. We observed that these methods achieved similar results in terms of the DSC, with scores of 77.20\%, 77.02\%, and 77.07\%, respectively. Our proposed SGPA-Net method, on the other hand, made use of point annotation. The results demonstrated that SGPA-Net performed exceptionally well in terms of DSC, reaching 79.39\%, and also exhibited strong performance in D$_{IRF}$ and D$_{PED}$, with scores of 71.13\% and 81.75\%, respectively. Fig.~\ref{fig6} showcases the outcomes obtained by SGPA-Net in comparison to other SSL approaches. The visual evidence highlights the remarkable alignment between the mask predicted by our method and the ground truth, providing further compelling validation of the efficacy of SGPA-Net.

In addition, we also investigated label-denoising methods, which enhance fluid segmentation performance by further optimizing existing noisy pseudo-labels.
In the label-denoising subgroup, we examined various methods, including Co-teaching, TriNet, 2SRnT, MTCL, and Dast. The results demonstrated outstanding performance across different metrics, with our proposed SCLGPA-Net achieving 80.89\% in DSC, 73.16\% in D$_{IRF}$, and 84.08\% in D$_{PED}$. This underscores the potential of label-denoising methods in further enhancing fluid segmentation performance. Fig.~\ref{fig7} presents the results achieved by SCLGPA-Net in contrast to alternative label-denoising methods. The visual evidence vividly illustrates the striking correspondence between our method's predicted mask and the ground truth, further substantiating the effectiveness of SCLGPA-Net.
\subsubsection{Comparative Experiments with Other Semi-Supervised Methods for OCT Fluid Segmentation}
In Tab~\ref{tab4}, we present a performance comparison on the RETOUCH dataset for various methods. The table evaluates the mIoU at different numbers of annotations ($N_l$) and reports the corresponding standard deviations.

Firstly, we observe the performance of the baseline (U-Net) method at different annotation counts. Our baseline model performs the worst at $N_l$=3, with a mIoU of 0.15, gradually improving as the number of annotations increases, reaching a mIoU of 0.49 at $N_l$=24.

Comparatively, semi-supervised methods that introduce additional unlabeled data achieve better performance. For instance, the IIC method \cite{ji2019invariant} consistently outperforms the baseline at various annotation counts. Similarly, the methods proposed by Perone and Cohen-Add \cite{perone2019unsupervised} and MLDS \cite{reiss2021every} exhibit significant improvements, particularly at $N_l$=24. The RPG method \cite{seibold2022reference} and RPG$^+$ method \cite{seibold2022reference} also achieve notable performance at $N_l$=24.
However, the proposed SGPA-Net and SCLGPA-Net achieve comparable or even superior performance on the RETOUCH dataset compared to these semi-supervised methods. 

Particularly, SCLGPA-Net demonstrates the best performance, achieving mIoU values of 0.520 and 0.549 at $N_l$=3 and $N_l$=6, respectively, showcasing a substantial lead over other methods. This indicates that in scenarios with limited labeled data, the pseudo-labels generated by our approach can serve as full labeled data, significantly enhancing the network's performance. Furthermore, SCLGPA-Net attains mIoU values of 0.563 and 0.604 at $N_l$=12 and $N_l$=24, respectively, outperforming other methods to a certain extent and the small standard deviation of our method demonstrates the stability of our algorithm.
This suggests that in scenarios with relatively sufficient labeled data, the pseudo-labels generated by our approach can provide the network with additional knowledge, leading to improved performance.

\subsection{Analytical Ablation Study}
To verify the effectiveness of each component, we propose different variants to perform ablation studies on private datasets. Table~\ref{tab5}, Table~\ref{tab6}, and Table~\ref{tab7} shows our ablation experiments. Our ablation experiments are performed on the FD accounted for 30$\%$, and PD accounted for 70$\%$.

Table~\ref{tab5} demonstrates the impact of $\delta $ and MT architecture on the performance of the SCLGPA-Net, where $\delta$ refers to the set of label trust maps and MT denotes the Mean-Teacher architecture. Without the MT architecture, the model's average dice score decreased by $2.5\%$ and performed worse than the SGPA-Net. It shows that the network trained on MT architecture can effectively utilize the pure image information of PD to improve performance. Furthermore, Table~\ref{tab5} indicates that the label trust information contained in the label trust map can help the network avoid over-fitting noise. The average dice score of the final model's result was even less than $80\%$ when trained without the label trust map. In the Refinement stage, trust estimation error maps (set to 1) can improve the performance of the network compared to discarding the estimation error labels (set to 0) or leaving them static. These experimental results show that the components of the SCLGPA-Net can effectively mitigate the negative effects of pseudo-labels on the network for OCT fluid segmentation.

Table~\ref{tab6} displays the impact of different hyper-parameters $\alpha$ and $\beta$ of the loss function (Equation~\ref{loss}) on the SCLGPA-Net. As we already have the label trust map to constrain the loss of PD, we chose to set $\alpha = 1$ and $\beta = 1$, which performed optimally in terms of most metrics. SCLGPA-Net with appropriate hyper-parameters achieved superior results.

As the bulk of the training data consists of pseudo-labels generated by point annotations, conducting ablation studies on the SGPLG method to yield the best point-annotated data is critical. We set the superpixel block size to 13 in our experiment, meaning each superpixel block encompasses approximately 169 pixels (13 × 13 on average). A pivotal factor to consider is the setting of the similarity threshold. If the threshold for creating pseudo-labels is excessively high, the labels may convey insufficient information, leading to network under-fitting. Conversely, if the threshold is too low, the pseudo-labels could introduce an overwhelming amount of incorrect information, adversely affecting network performance. Therefore, careful selection of an appropriate threshold for generating pseudo-labels is essential to strike a balance between valid information and misinformation in the labels for optimum network performance. Compared to SRF and IRF, we noted that visually distinguishing PED fluids can be more challenging. Therefore, we set a smaller similarity threshold for PED when determining the threshold. Table~\ref{tab7} presents the final impact of our generated pseudo-labels for network training under different similarity thresholds. We selected $SI_t = 0.6, P_t = 0.5$ due to its superior performance across most metrics.

\begin{table}[!t]
\centering
    \renewcommand\arraystretch{1.5}
    \caption{Ablation Study On Private Dataset of different $\delta$ and whether to use MT architecture. The Best Results Are In Bold. (Dice Unit: $\%$)}
    \label{tab5}
    \large
\scalebox{0.7}{
\begin{tabular}{c|ll|llll}
\toprule[1pt]\hline
\multirow{2}{*}{\textbf{Methods}} & \multicolumn{2}{c|}{\textbf{Settings}}                              & \multicolumn{4}{c}{\textbf{Metrics}}                                                                                                                                                                                   \\ \cline{2-7} 
\multicolumn{1}{c|}{}                         & \multicolumn{1}{c|}{$\delta$}         & \multicolumn{1}{c|}{MT}   & \multicolumn{1}{c|}{DSC}      & \multicolumn{1}{c|}{D$_{SRF}$}      & \multicolumn{1}{c|}{D$_{IRF}$}      & \multicolumn{1}{c}{D$_{PED}$}            \\ \hline \hline
SGPA-Net                                 & \multicolumn{1}{c|}{-}         & \multicolumn{1}{c|}{-}    & \multicolumn{1}{c|}{81.24}          & \multicolumn{1}{c|}{87.91} & \multicolumn{1}{c|}{86.66}          & \multicolumn{1}{c}{69.16}            \\ \hline \hline
\multirow{5}{*}{SCLGPA-Net}                                     
& \multicolumn{1}{c|}{set to 1}     & \multicolumn{1}{c|}{$\checkmark$} & \multicolumn{1}{c|}{\textbf{82.87}}          & \multicolumn{1}{c|}{\textbf{89.46}}          & \multicolumn{1}{c|}{\textbf{88.78}} & \multicolumn{1}{c}{70.37}  \\
                                       & \multicolumn{1}{c|}{set to 1}     & \multicolumn{1}{c|}{$\times$}                         & \multicolumn{1}{c|}{80.17}          & \multicolumn{1}{c|}{86.87}          & \multicolumn{1}{c|}{86.59}          & \multicolumn{1}{c}{67.04}                   \\ 
                                      & \multicolumn{1}{c|}{static} & \multicolumn{1}{c|}{$\checkmark$} & \multicolumn{1}{c|}{82.14} & \multicolumn{1}{c|}{88.41}          & \multicolumn{1}{c|}{86.45}          & \multicolumn{1}{c}{\textbf{71.54}}            \\
                                       & \multicolumn{1}{c|}{set to 0}     & \multicolumn{1}{c|}{$\checkmark$} & \multicolumn{1}{c|}{81.63}          & \multicolumn{1}{c|}{89.03}          & \multicolumn{1}{c|}{87.28}          & \multicolumn{1}{c}{68.59}              \\
                                       & \multicolumn{1}{c|}{$\times$}         & \multicolumn{1}{c|}{$\checkmark$} & \multicolumn{1}{c|}{79.73}          & \multicolumn{1}{c|}{85.08}          & \multicolumn{1}{c|}{84.80}          & \multicolumn{1}{c}{69.29}          \\ \hline \bottomrule[1pt]
\end{tabular}}
\end{table}

\begin{table}[!t]
\centering
    \renewcommand\arraystretch{1.5}
    \caption{Ablation Study Of Different Loss Weight $\beta$ Of PD on Private Dataset. The Best Results Are In Bold.  (Dice Unit: $\%$)}
    \label{tab6}
    \large
\scalebox{0.75}{
\begin{tabular}{c|cc|cccc}
\toprule[1pt] \hline
\multirow{2}{*}{\textbf{Methods}} & \multicolumn{2}{c|}{\textbf{Settings}}                                                          & \multicolumn{4}{c}{\textbf{Metrics}}                                                                                                                                                                     \\ \cline{2-7} 
\multicolumn{1}{c|}{}                         & \multicolumn{1}{c|}{$\alpha$} & \multicolumn{1}{c|}{$\beta$} & \multicolumn{1}{c|}{DSC} & \multicolumn{1}{c|}{D$_{SRF}$}      & \multicolumn{1}{c|}{D$_{IRF}$}      & \multicolumn{1}{c}{D$_{PED}$}  \\ \hline \hline
\multirow{3}{*}{SCLGPA-Net}                          
 
& \multicolumn{1}{c|}{1}                     & 0.3                                       & \multicolumn{1}{c|}{81.57}     & \multicolumn{1}{c|}{86.10}          & \multicolumn{1}{c|}{85.59}          & \multicolumn{1}{c}{\textbf{72.72}}    \\ 
                                       & \multicolumn{1}{c|}{1}                   & 0.5                                         & \multicolumn{1}{c|}{80.99}     & \multicolumn{1}{c|}{87.60}          & \multicolumn{1}{c|}{83.79}          & \multicolumn{1}{c}{71.57}           \\ 
                                       & \multicolumn{1}{c|}{1}                     & \multicolumn{1}{c|}{1}       & \multicolumn{1}{c|}{\textbf{82.87}}     & \multicolumn{1}{c|}{\textbf{89.46}} & \multicolumn{1}{c|}{\textbf{88.78}} & \multicolumn{1}{c}{70.37}  \\ \hline \bottomrule[1pt]

\end{tabular}}
\end{table}

\begin{table}[!t]
\centering
\small
    \renewcommand\arraystretch{1.5}
    \caption{Ablation Study Of Different $t$ Of PED, IRF, And SRF when Generating Pseudo-Label By SGPLG on Private Dataset. The Best Results Are In Bold. (Dice Unit: $\%$)}
    \label{tab7}
    \large
\scalebox{0.76}{
\begin{tabular}{c|ll|llll}
\toprule[1pt]
\hline
\multirow{2}{*}{\textbf{Methods}} & \multicolumn{2}{c|}{\textbf{Settings}}              & \multicolumn{4}{c}{\textbf{Metrics}}                                                                                                                                                                                            \\ \cline{2-7} 
\multicolumn{1}{c|}{}                         & \multicolumn{1}{c|}{SI$_t$} &P$_t$&   \multicolumn{1}{c|}{DSC} & \multicolumn{1}{c|}{D$_{SRF}$} & \multicolumn{1}{c|}{D$_{IRF}$} & \multicolumn{1}{c}{D$_{PED}$}    \\ \hline \hline
\multirow{3}{*}{SCLGPA-Net}
& \multicolumn{1}{c|}{0.7}    & 0.6         & \multicolumn{1}{c|}{79.35}            & \multicolumn{1}{c|}{83.51}            & \multicolumn{1}{c|}{86.54}            & \multicolumn{1}{c}{69.19}                 \\
                                       & \multicolumn{1}{c|}{0.6}    & 0.5         & \multicolumn{1}{c|}{\textbf{82.87}}   & \multicolumn{1}{c|}{89.46}   & \multicolumn{1}{c|}{\textbf{88.78}}            & \multicolumn{1}{c}{\textbf{70.37}}        \\

                                       & \multicolumn{1}{c|}{0.5}    & 0.4         & \multicolumn{1}{c|}{81.52}            & \multicolumn{1}{c|}{\textbf{90.87}}            & \multicolumn{1}{c|}{83.55}            & \multicolumn{1}{c}{70.16}                \\ \hline \bottomrule[1pt]
\end{tabular}}
\end{table}

\section{Discussion}

Recently, CNN-based segmentation methods have achieved tremendous success in various applications. However, the training of CNN-based segmentation models heavily relies on pixel-wise manual segmentation masks and has limited performance improvement with the use of unlabeled data in semi-supervised approaches. Our method only requires simple point annotation on the unlabeled data, reducing the need for precise annotations while significantly improving model performance. Specifically, we proposed SGPLG and CLGLR modules for reliable learning from abundant PD. The SGPLG can generate pseudo-labels with weights based on point annotation. To mitigate the negative impact of imprecise pseudo-labels on model performance, the SGPLG can identify labeling errors in pseudo-labels and perform further refinement to obtain more accurate labels.

The SGPLG module is heavily dependent on the similarity among superpixel blocks, leading to the generation of pseudo-labels with noticeable gaps. These gaps can substantially impact the training process of the model. To rectify this, we utilized the CLGLR module to mend the gaps with reliable guidance. The refined-labels obtained by CLGLR are more closely aligned with the ground truths compared to the original pseudo-labels. The CLGLR module fills in the gaps and refines the edges of the labels, resulting in notable improvements. The superior effectiveness of our label-denoising process is highlighted both in the final performance of the model and in visual depictions of the label-denoising process. As evidenced by the results in Fig.~\ref{fig8}, the CLGLR module exhibits impressive effectiveness.

While our method has shown promising results, it is important to note that it still relies on a moderate quantity of fully-annotated data. As a forward-looking research direction, we aspire to mitigate the impact of this distribution difference on model training. One approach on our radar involves the incorporation of data augmentation 
techniques, such as Cutmix~\cite{yun2019cutmix}, to further enhance the model's ability to generalize across diverse data distributions. This will ultimately contribute to the development of more efficient and reliable medical image segmentation models, which can significantly aid in the diagnosis and treatment of various medical conditions.

\section{Conclusion}
In this study, we introduced the Superpixel and Confident Learning Guide Point Annotations Network (SCLGPA-Net), which is built on a teacher-student architecture. The innovative approach enables the learning of OCT fluid segmentation from both limited fully-annotated data and abundant point-annotated data. 
Compared to the additional unlabeled data introduced by semi-supervised methods, point annotations contain more valuable information and can significantly enhance model performance.
To make full and accurate use of point annotation information, our method incorporates two key modules: the Superpixel-Guided Pseudo-Label Generation (SGPLG) module, which generates pseudo-labels with weights from point annotations; and the Confident Learning Guided Label Refinement (CLGLR) module, designed to identify errors in the pseudo-labels and further refine them.
We evaluated the effectiveness of our approach using a private 2D OCT fluid segmentation dataset and a public RETOUCH dataset. The results of extensive experiments on the OCT fluid segmentation dataset show that our method performs well. The visualization results further validate the excellent performance and our method exhibits better segmentation results for OCT image fluid compared to other methods, highlighting our ability to capture segmentation boundaries and generate better masks accurately. 

\section{Acknowledge}
This research was supported by the Zhejiang Provincial Natural Science Foundation of China under Grant No. LY21F020004, National Natural Science Foundation of China under Grants No. 62201323 and No. 62206242, and Natural Science Foundation of Jiangsu Province under Grant No. BK20220266.
\bibliographystyle{IEEEtran}
\bibliography{IEEEabrv,myrefs}

\end{document}